\documentclass[conference]{IEEEtran}
\IEEEoverridecommandlockouts
\usepackage{cite}
\usepackage{amsmath,amssymb,amsfonts}
\usepackage{algorithmic}
\usepackage{graphicx}
\usepackage{textcomp}
\usepackage{xcolor}
\usepackage{subcaption}
\usepackage{booktabs}
\usepackage{dirtytalk}

\def\BibTeX{{\rm B\kern-.05em{\sc i\kern-.025em b}\kern-.08em
    T\kern-.1667em\lower.7ex\hbox{E}\kern-.125emX}}
\begin{document}

\title{BlenDAgger: Blended Shared Control for Interactive Imitation Learning
}

\author{\IEEEauthorblockN{Cailyn Smith}
\IEEEauthorblockA{\textit{The Robotics Institute,} \\
\textit{School of Computer Science}\\
\textit{Carnegie Mellon University}\\
Pittsburgh, USA \\
cailyns@andrew.cmu.edu}
\and
\IEEEauthorblockN{Geoffrey Sun}
\IEEEauthorblockA{\textit{School of Computer Science}\\
\textit{Carnegie Mellon University}\\
Pittsburgh, USA \\
gsun2@andrew.cmu.edu}
\and
\IEEEauthorblockN{Henny Admoni\textsuperscript{\dag}}
\IEEEauthorblockA{\textit{The Robotics Institute,} \\
\textit{School of Computer Science}\\
\textit{Carnegie Mellon University}\\
Pittsburgh, USA \\
hadmoni@andrew.cmu.edu }
\and
\IEEEauthorblockN{Zackory Erickson\textsuperscript{\dag}}
\IEEEauthorblockA{\textit{The Robotics Institute,} \\
\textit{School of Computer Science}\\
\textit{Carnegie Mellon University}\\
Pittsburgh, USA \\
zackory@cmu.edu}
}
\maketitle

\begingroup
\renewcommand\thefootnote{\dag}
\footnotetext{Equal advising.}
\endgroup

\begin{abstract}
Robot policies are frequently trained from human corrections, yet teleoperating a robot to provide corrections is burdensome, and human demonstrators are not always optimal. We propose Blended DAgger (BlenDAgger), an approach for collecting data to train imitation learning policies by using shared control to blend the policy's and demonstrator's actions during interventions. By blending human and policy actions, we aim to improve the autonomous performance of manipulation policies. We validate our approach across five manipulation tasks, two in the real world and three in simulation. Our approach achieves higher autonomous performance by 30 or more percentage points on two real-world tasks compared to a typical human-gated correction approach (HG-DAgger). We also investigate the advantages of BlenDAgger that allow for higher autonomous performance, finding that BlenDAgger results in 57\% smoother transitions between policy control and human interventions, and 14\% higher trajectory similarity to the training data. In a user study (n=14) on two real-world tasks, we find that BlenDAgger results in faster data collection (BF=13.32), and we do not find a difference in subjective perceptions. These results show that blended shared control leads to higher autonomous performance compared to typical methods for fine-tuning robot policies from fully teleoperated interventions.


\end{abstract}

\begin{IEEEkeywords} Imitation learning, telerobotics and teleoperation, human factors and human-in-the-loop
\end{IEEEkeywords}

\section{Introduction}
\label{sec:intro}
\begin{figure}[t]
\centering
\includegraphics[width=\columnwidth]{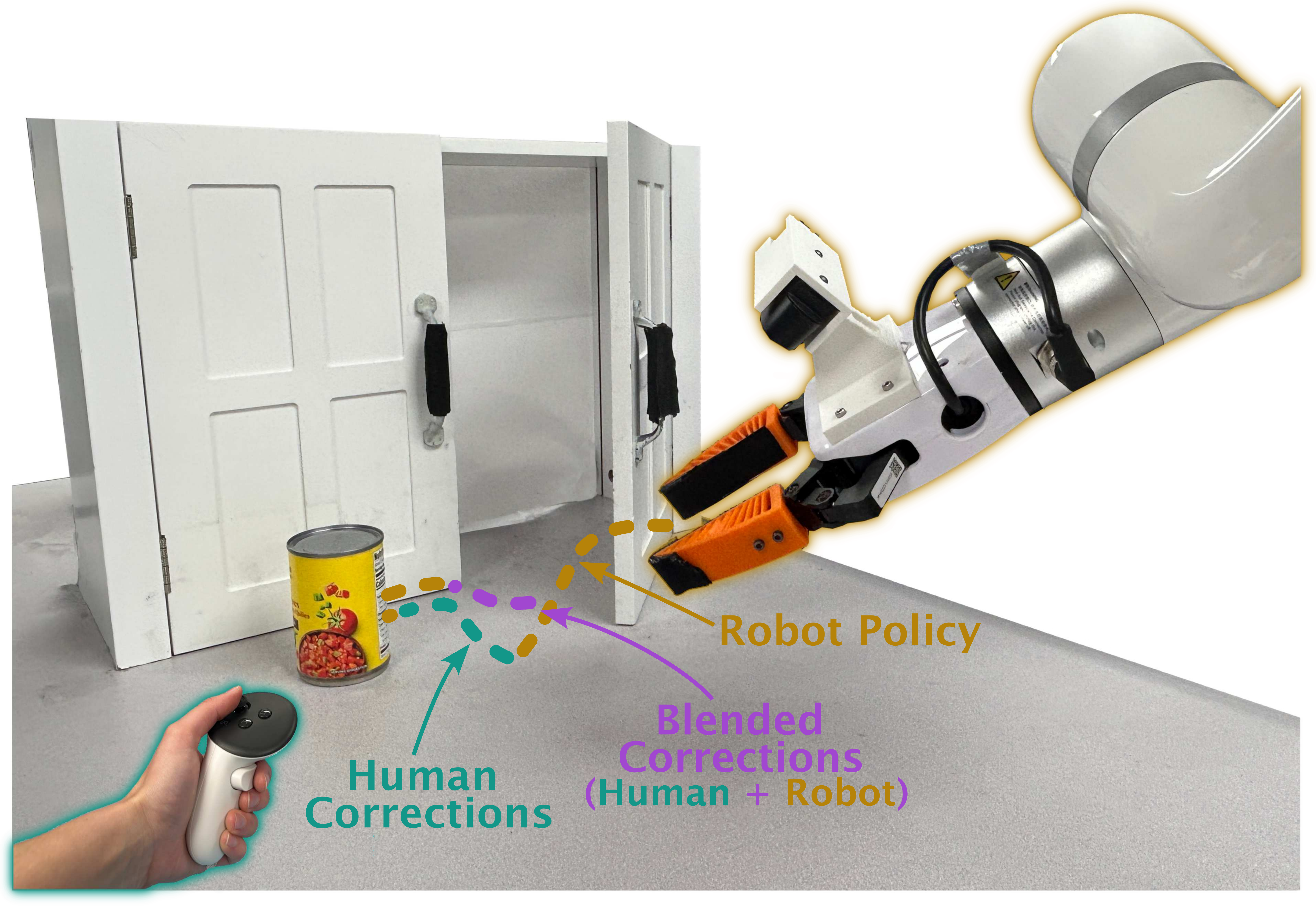}
\caption{BlenDAgger shares control between the robot's policy and the human demonstrator during corrections, resulting in higher autonomous success and faster data collection.}
\label{fig:teaser}
\end{figure}

Imitation learning has been successfully used to train policies to accomplish complex manipulation tasks using human-collected demonstration data~\cite{Zhao_Aloha_2023, Wu_Tidybot_2024, Fu_MobileALOHA_2024, Wu_RoboCopilot}.
To address the covariate shift introduced by the sequential nature of manipulation tasks, researchers often use DAgger~\cite{Ross_DAgger_2011} and its variants, in which an expert corrects policy errors during execution. Human-gated and robot-gated forms of DAgger~\cite{Kelly_HG-DAgger, Spencer_EIL, Menda_Ensemble_2019} have shown success in alleviating the covariate shift problem. However, these gated interventions put the human fully in control during interventions. Fully teleoperating a robot can result in suboptimal interventions and may be burdensome. Our insight is that by using \textit{blended} shared control, where the human's and robot's actions are combined at each timestep based on their similarity and the robot's uncertainty, we can improve the autonomous performance of manipulation policies (see Fig.~\ref{fig:teaser}).

Blended shared control~\cite{Dragan_PolicyBlending} has been studied extensively in assistive robotics~\cite{Gopinath_SharedControl_2016, Javdani_SA-hindsight, Losey_SharedControlReview_2018}. Some existing work incorporates shared control to expand the robot's skills as the user interacts with the system multiple times~\cite{Tao_ILSA, Jonnavittula_SARI, Zurek_CASA}, but these works focus on shared control as a way to improve human-robot collaboration where the human remains in the loop. Our work aims to do the reverse: rather than using imitation learning to improve shared control systems, we leverage blended shared control to fine-tune manipulation policies. We ask the research question: \textit{Can blended shared control provide a valuable learning signal for imitation learning?}

Blending the human's and robot's actions at each timestep offers several potential benefits over gated DAgger approaches:
\begin{enumerate}
    \item \textbf{Smoother corrections.} Combining suboptimal human inputs with policy actions may smooth out corrections.
    \item \textbf{Reduced excursions to out-of-distribution (OOD) states.} Unlike gated control, which alternates between the policy's OOD rollout states and human-corrected states, blending may keep the robot more consistently near in-distribution states.
    \item \textbf{Better-timed interventions.} Seeing the policy's actions while correcting may help demonstrators judge when to start and stop intervening.
\end{enumerate}


We evaluate BlenDAgger for fine-tuning policies on five manipulation tasks: two in the real world and three in simulation. We find that BlenDAgger achieves higher autonomous task performance than HG-DAgger on four tasks when trained on data collected by experienced demonstrators, with 30 or more percentage points higher performance on the real-world tasks. To evaluate whether the autonomous performance gains of our approach extend to non-expert demonstrators, we conduct a user study with 14 novice users. We find that BlenDAgger reduces task time and do not find an impact on subjective perceptions of providing corrections. We further find that BlenDAgger outperforms HG-DAgger on three of the four conditions when trained with participant data, but that gains from both BlenDAgger and HG-DAgger are smaller than with expert data.

We make the following contributions:
\begin{enumerate}
    \item We propose an interactive imitation learning framework, Blended DAgger (\emph{BlenDAgger}), that integrates adaptive, uncertainty-aware shared control into policy fine-tuning.
    \item We demonstrate the efficacy of BlenDAgger across diverse simulation and real-world manipulation tasks, showing substantial gains in policy success rates over a traditional gated baseline.
    \item We conduct a user study with novice operators to evaluate our proposed method against HG-DAgger, providing insights into blended shared control for data collection.
\end{enumerate}

\section{Related Work}
\subsection{Human- and Robot-Gated DAgger}
Existing approaches for learning from human interventions are typically either human-gated or robot-gated.
In human-gated approaches, such as HG-DAgger~\cite{Kelly_HG-DAgger} and EIL~\cite{Spencer_EIL}, a human supervises the policy rollouts and intervenes as needed to correct the policy. HG-DAgger has been extended in many works, including by modifying its data sampling during training~\cite{Mandlekar_IWR_2020, Liu_Learning-on-the-job}, learning a residual policy~\cite{Jiang_Transic_2024}, or using a compliant interface~\cite{Xu_Compliant_2025}. In robot-gated approaches~\cite{Hoque_LazyDAgger_2021, Hoque_ThriftyDAgger_2022, Menda_Ensemble_2019}, the robot actively queries an expert when unsure of the best action or in a risky state. In contrast, our approach is not gated but instead allows the human demonstrator to provide corrections that are blended with policy actions.
\subsection{Bilateral and Compliant Control}
Existing work has incorporated ideas of continuous shared control with imitation learning, but not with explicit blending of policy and human actions. Work on bilateral control, including HACTS~\cite{Xu_HACTS_2025} and RoboCopilot~\cite{Wu_RoboCopilot}, aims to synchronize autonomous policy rollouts with human interventions primarily through hardware advancements. CHG-DAgger~\cite{Takahashi_CHG} uses multilateral control to combine human interventions with policy rollouts rather than using a gated mechanism. Unlike our work, their use of multilateral control means that there is no explicit determination of shared autonomy arbitration at the action level. Furthermore, kinesthetic teaching and compliant interfaces have been used in prior work. By providing force corrections during policy rollouts rather than action labels, these interfaces, such as CR-DAgger~\cite{Xu_Compliant_2025} and that of Abi-Farraj et al.~\cite{Abi-Farraj_Kinesthetic-learning}, effectively combine human and robot actions during deployment through the force applied to the robot rather than by explicitly blending those actions with shared control.

\subsection{Learning from Shared Autonomy}
Prior works have used shared control for learning manipulation tasks, typically optimizing for the quality of shared control itself. Many of these works use shared control in a discrete, gated fashion rather than blending actions. ILSA~\cite{Tao_ILSA} uses a gated mechanism between human and robot control depending on how significantly their actions differ. In~\cite{Cui_VLA-SharedAutonomy}, Cui et al. incorporate shared autonomy where the human demonstrator has full control of the arm movements while the robot policy autonomously commands a dexterous hand. SARI~\cite{Jonnavittula_SARI} and CASA~\cite{Zurek_CASA} blend human and robot control while deferring to the human when the robot encounters an unfamiliar task or goal. These works aim to improve the robot's behavior to stay in a shared control loop and therefore do not compare learning efficiency to gated imitation learning approaches. Li et al.~\cite{Li_HandInLoop_2026} blend human corrections and policy actions for wrist motion in dexterous manipulation by adding human velocity commands to the policy velocity, which effectively results in a fixed arbitration value rather than an adaptive one. In contrast to prior work, our proposed algorithm leverages existing adaptive blended shared control formulations from assistive robotics, using action similarity and policy confidence, and combines this with imitation learning to fine-tune manipulation policies.



\section{Formulation}
\label{sec:formulation}

\begin{figure*}[t]
\centering
\includegraphics[width=0.95\textwidth]{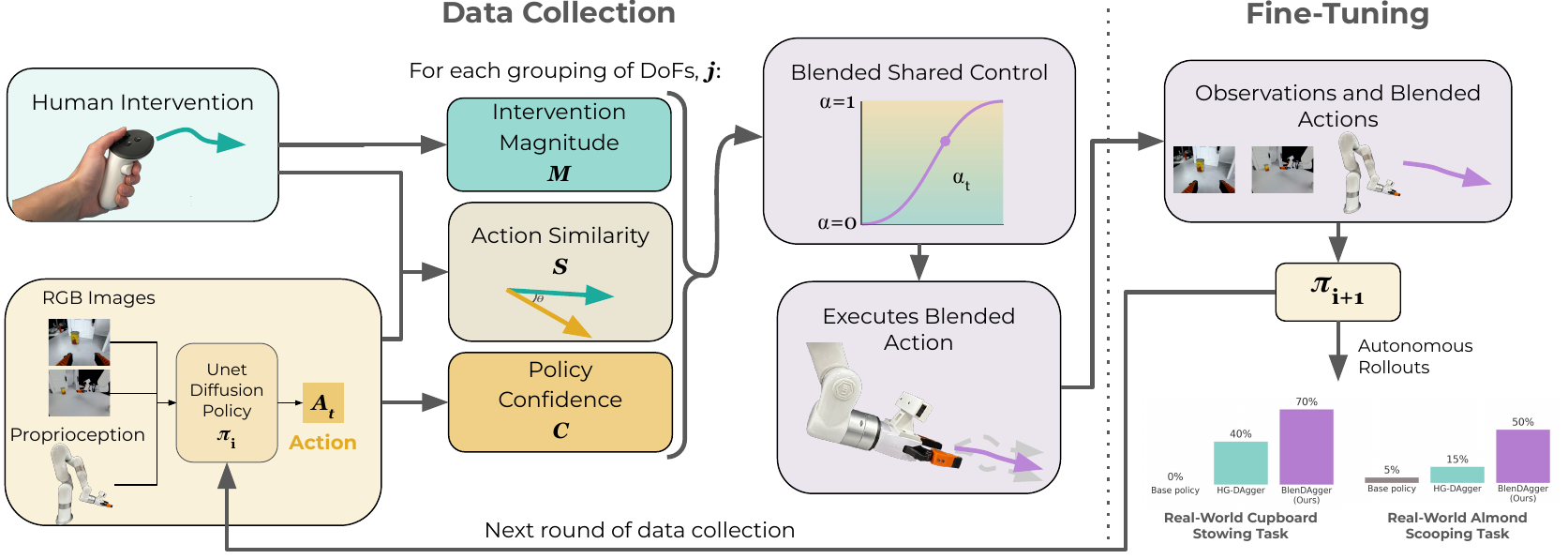}
\caption{BlenDAgger pipeline. When the demonstrator chooses to intervene, their action is blended with the policy's predicted action. The resulting blended action is executed on the robot and used to fine-tune the policy.}
\label{fig:pipeline}
\end{figure*}

\subsection{Preliminaries}
We train a robot policy $\pi_\theta$ on a demonstration dataset
$\mathcal{D} = \{(\mathbf{O}_t, \mathbf{A}_t)\}$, where each
$\mathbf{O}_t = (o_{t-T_o+1}, \dots, o_t)$ is a history of $T_o$ observations
$o \in \mathcal{O}$ containing RGB images and proprioceptive state, and
$\mathbf{A}_t = (a_t, \dots, a_{t+15})$ is an action chunk of length $16$.
Following Chi et al.~\cite{Chi_Diffusion_2024}, the policy models the
conditional distribution $\pi_\theta(\mathbf{A}_t \mid \mathbf{O}_t)$ and is trained
with the diffusion denoising objective. At inference, the policy samples an action chunk using Denoising Diffusion Implicit Models
(DDIM)~\cite{Song_DDIM_2020} and executes $8$ actions before replanning.

In HG-DAgger~\cite{Kelly_HG-DAgger}, the executed policy is gated on whether the
demonstrator intervenes:
\begin{equation}
  \pi_i(x_t) = g(x_t)\, \pi_H(x_t) + \big(1 - g(x_t)\big)\, \pi_{\theta_i}(\mathbf{O}_t),
  \label{eq:hgdagger_gate}
\end{equation}
where $i$ is the aggregation round, $\pi_H(x_t)$ is the demonstrator's policy, $x_t$ is the full state, and
$g(x_t)\in \{0,1\}$ is $1$ when the demonstrator chooses to intervene and is $0$ otherwise. Each round's data $\mathcal{D}_i$ is aggregated, $\mathcal{D} \leftarrow \mathcal{D} \cup \mathcal{D}_i$, before retraining.

\subsection{BlenDAgger}
In contrast to HG-DAgger, we leverage the policy blending formalism from~\cite{Dragan_PolicyBlending}, which is frequently used in goal-conditioned assistive robotics domains. Rather than gating the policy based on human interventions, we blend control as: 
\begin{equation}
    \pi_i(x_t) = (1-\alpha_t)\ \pi_H(x_t) + \alpha_t\ \pi_{\theta_i}(\mathbf{O}_t),
    \label{eq:blended}
\end{equation}
where $\alpha_t \in [0,1]$ is the arbitration function. Following the convention
from~\cite{Dragan_PolicyBlending}, $\alpha_t = 0$ corresponds to pure
teleoperation and
$\alpha_t = 1$ to full autonomy. This is a superset of HG-DAgger, becoming HG-DAgger if $\alpha$ goes to $0$ during interventions. Unlike traditional policy blending, which predicts the user's goal and assists toward it~\cite{Dragan_PolicyBlending}, we assume no goal set and blend only at the trajectory execution layer. 


\subsection{Arbitration Function}
\label{sec:arbitration}
\newcommand{\rvec}{\mathbf{a}^{r}}
\newcommand{\hvec}{\mathbf{a}^{h}}
When the human operator decides to intervene during policy execution, they use a device that allows 6DoF control. However, since operators rarely control all 6DoFs simultaneously, we group together DoFs and then compute $\alpha$ separately for each DoF group. Letting $x,y,z$ denote translation and $\phi, \theta, \psi$ denote roll, pitch, and yaw, the DoF groups are,

\begin{equation}
  j \in \big\{\, \{x, y\},\ \{z\},\ \{\phi, \theta, \psi\} \,\big\}.
  \label{eq:dof_groups}
\end{equation}
We chose these groupings since manipulation tasks are often done on planar, human-made surfaces. We use $\rvec_{j,t}$ and $\hvec_{j,t}$ for the robot and human commands at
timestep $t$ of group $j$, with each axis normalized to
$[-1,1]$.

Taking inspiration from shared control systems in assistive robotics that
frequently blend control based on confidence~\cite{Dragan_PolicyBlending, Gopinath_SharedControl_2016, Javdani_SA-hindsight} and similarity~\cite{Abou_SharedControl_2019, Collier_SoA_2025}, we utilize the
following components as part of blending arbitration (see Fig.~\ref{fig:pipeline}):

\begin{itemize}
    \item \textbf{Similarity.}
    We use cosine similarity between the policy-predicted action and the human action:
    \begin{equation}
      S_{j,t} = \frac{(\hvec_{j,t})^{T} \rvec_{j,t}}
                     {\lVert \hvec_{j,t} \rVert \, \lVert \rvec_{j,t} \rVert}.
    \label{eq:similarity}
    \end{equation}
    Since we use delta actions and represent orientation in axis-angle form, the action space is approximately linear, which satisfies the assumption behind cosine similarity.

    \item \textbf{Policy uncertainty.}
    We estimate policy uncertainty with $k$-nearest neighbors~\cite{Sun_KNN_2022}, using the Euclidean distance from the policy's encoder embedding (see Section~\ref{sec:arch})
    of the current observation to the $10$-th nearest embedding of the
    policy's training data. This is normalized to $\hat{u}_t \in [0,1]$ using bounds calibrated on that same data, which yields a confidence $C_t = 1 - \hat{u}_t$. We recalibrate uncertainty with each data aggregation round.

    \item \textbf{Magnitude of intervention.} As a demonstrator pushes more in a specific direction, this may indicate that they want a quicker correction in that direction. We compute this as the magnitude of the normalized human action,
    \begin{equation}
      M_{j,t} = \min\!\big( \lVert \hvec_{j,t} \rVert,\ 1 \big).
    \end{equation}
\end{itemize}

\begin{figure*}[ht!]
\centering
\begin{subfigure}[a]{0.19\textwidth}
    \centering
    \includegraphics[width=\textwidth]{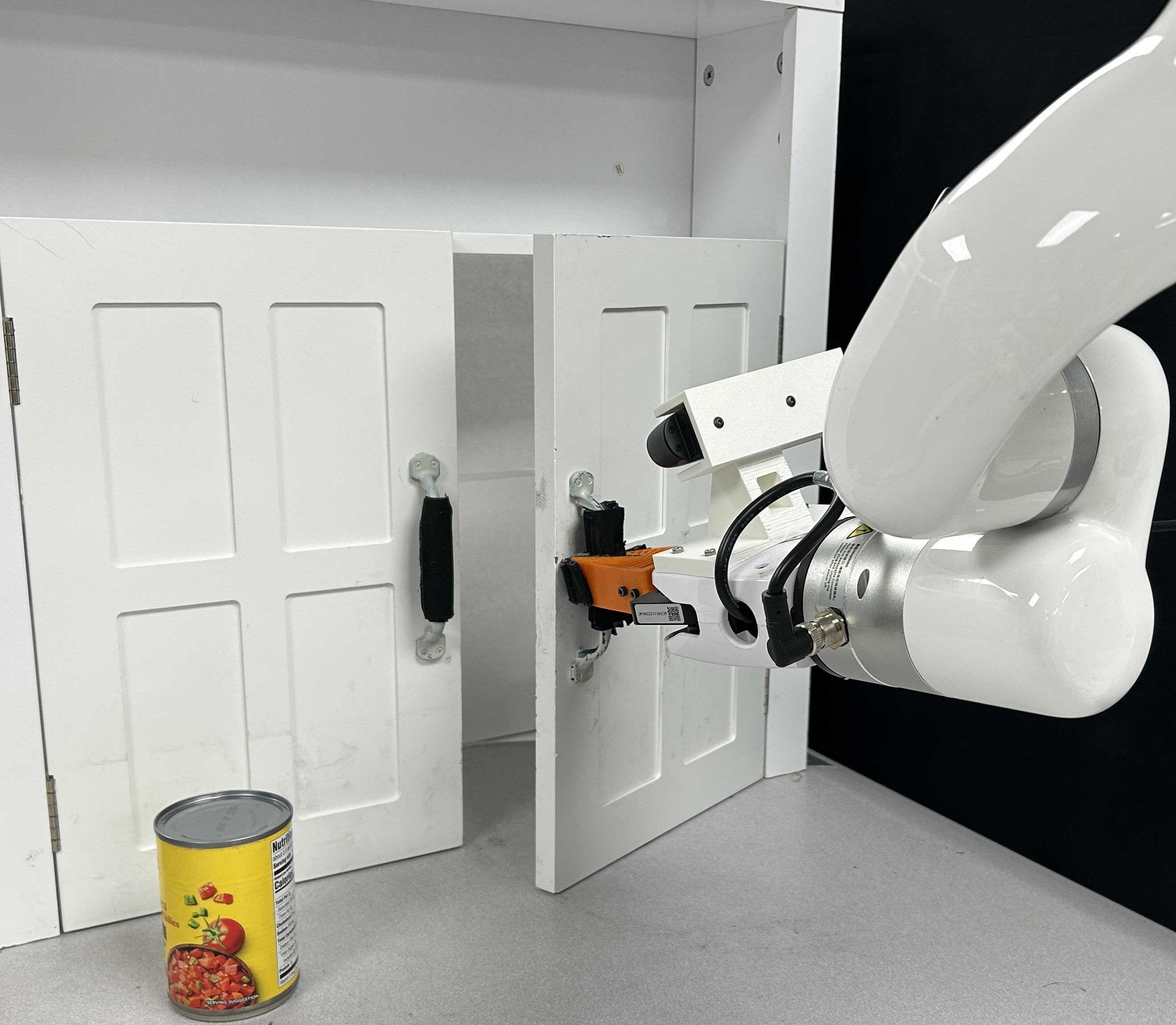}
    \caption{Cupboard Stowing}
    \label{fig:cupboard-task}
\end{subfigure}
\hfill
\begin{subfigure}[a]{0.19\textwidth}
    \centering
    \includegraphics[width=\textwidth]{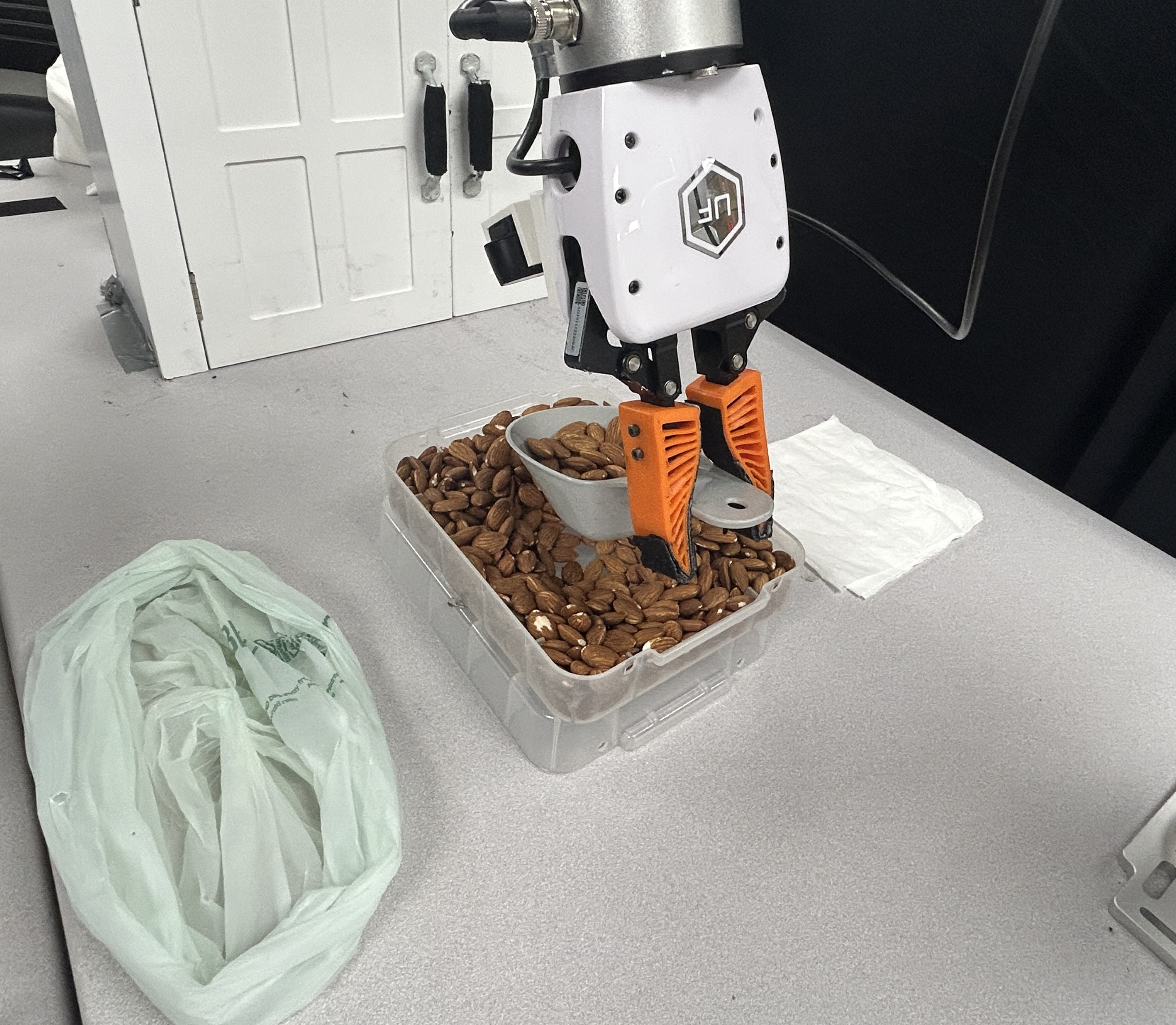}
    \caption{Almond Scooping}
    \label{fig:almond-task}
\end{subfigure}
\hfill
\begin{subfigure}[a]{0.19\textwidth}
    \centering
    \includegraphics[width=\textwidth]{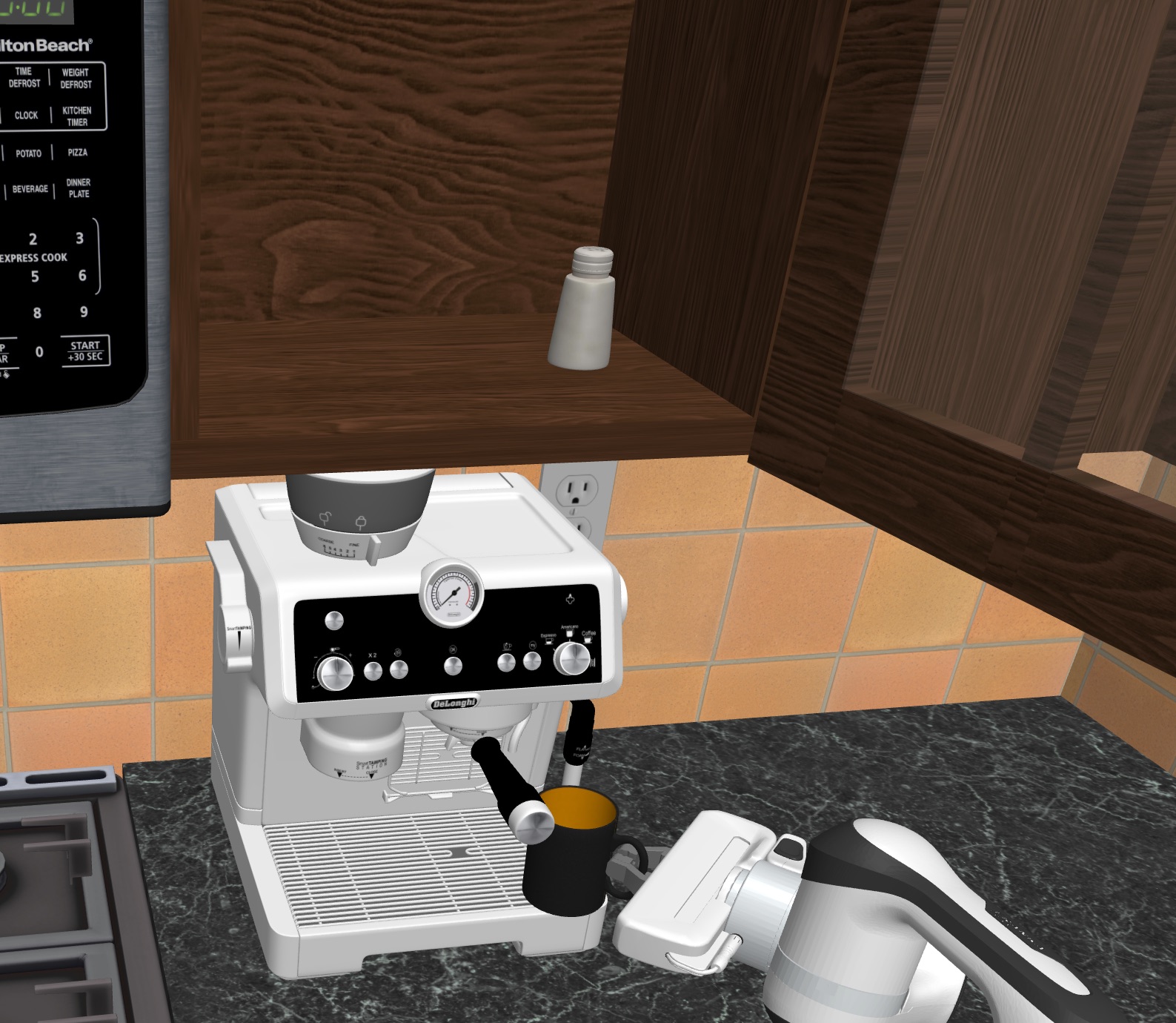}
    \caption{Prepare Coffee}
    \label{fig:coffee-task}
\end{subfigure}
\hfill
\begin{subfigure}[a]{0.19\textwidth}
    \centering
    \includegraphics[width=\textwidth]{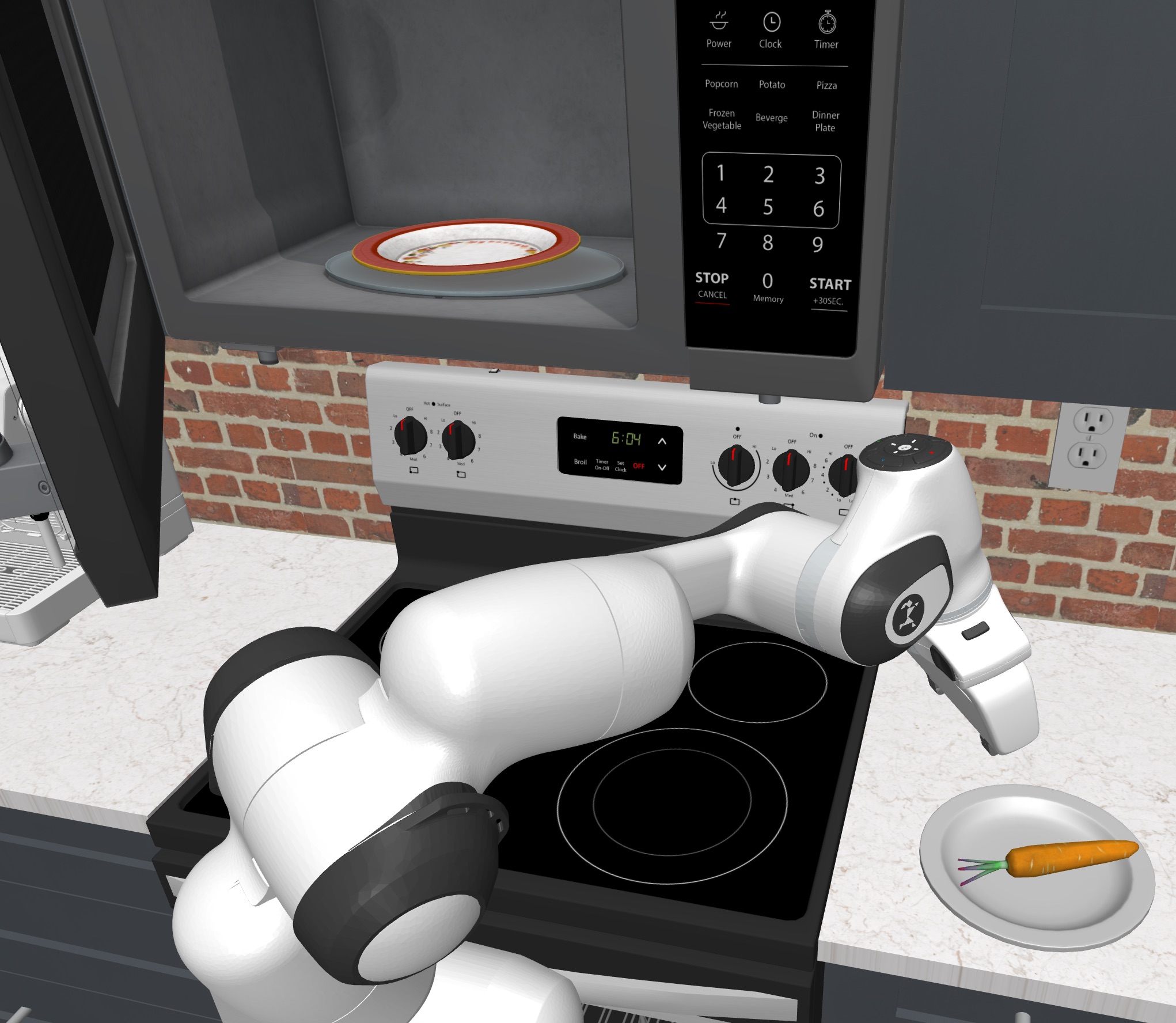}
    \caption{Microwave Thawing}
    \label{fig:thawing-task}
\end{subfigure}
\hfill
\begin{subfigure}[a]{0.19\textwidth}
    \centering
    \includegraphics[width=\textwidth]{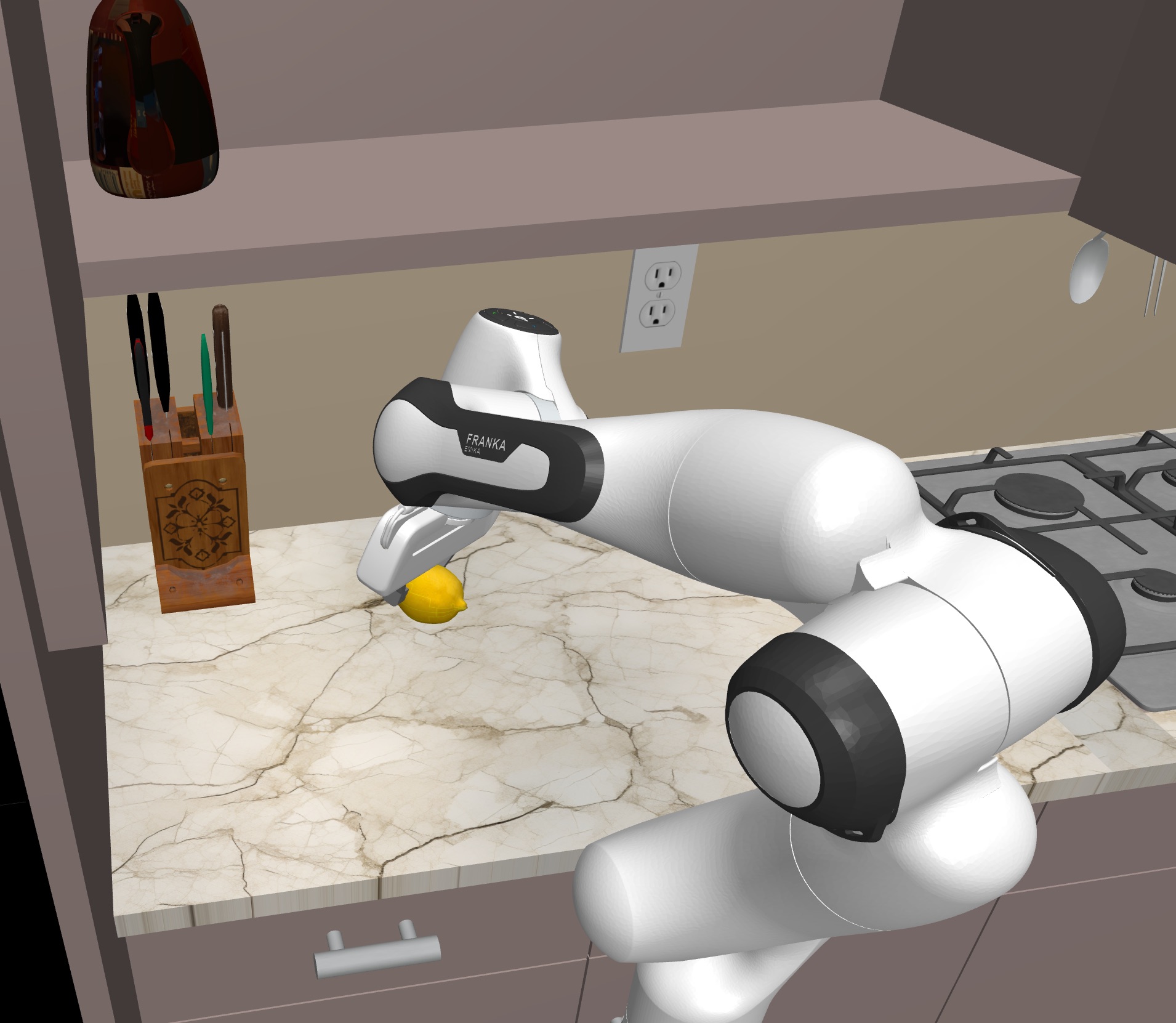}
    \caption{Pick and Place}
    \label{fig:lemon-task}
\end{subfigure}
\caption{Experiment Tasks.}
\label{fig:tasks}
\end{figure*}

Using these components, we compute $\alpha$ with a sigmoid, as sigmoids and other ramping functions have often been used in the shared control literature~\cite{Muelling_SharedControl_2015, Dragan_PolicyBlending, Gopinath_SharedControl_2016}. Action similarity and policy confidence are each passed through a sigmoid:
\begin{equation}
  f(w, M) = \frac{1}{1 + \exp\!\big( -\beta \cdot (w - M - \delta) \big)},
  \label{eq:sigmoid}
\end{equation}
where $\beta = 10$ is a fixed slope and $\delta = 0.5$ centers the
function on $[0,1]$. When the demonstrator actively intervenes to control group $j$, the target arbitration for that group is the sigmoid resulting from the product of the two terms:
\begin{equation}
  \alpha^{\text{tgt}}_{j,t} = f\big(S_{j,t},\, M_{j,t}\big) \cdot f\big(C_t,\, M_{j,t}\big).
  \label{eq:alpha_target}
\end{equation}
This arbitration function allows for more human control when the demonstrator's actions diverge from the policy's predicted actions, the policy has high uncertainty, or the demonstrator's input magnitude is large. This allows for blending while still providing the operator with sufficient ways to regain control (see Section~\ref{sec:ablations} for analysis of responsiveness). Finally, the arbitration is smoothed across timesteps so that the demonstrator is able to react to the changing amount of control:
\begin{equation}
  \alpha_{j,t} = \lambda\, \alpha_{j,t-1} + (1 - \lambda)\, \alpha^{\text{tgt}}_{j,t}\ ;\ \lambda=0.5.
  \label{eq:alpha_ema}
\end{equation}

\begin{figure}[b]
\centering
\includegraphics[width=0.47\textwidth]{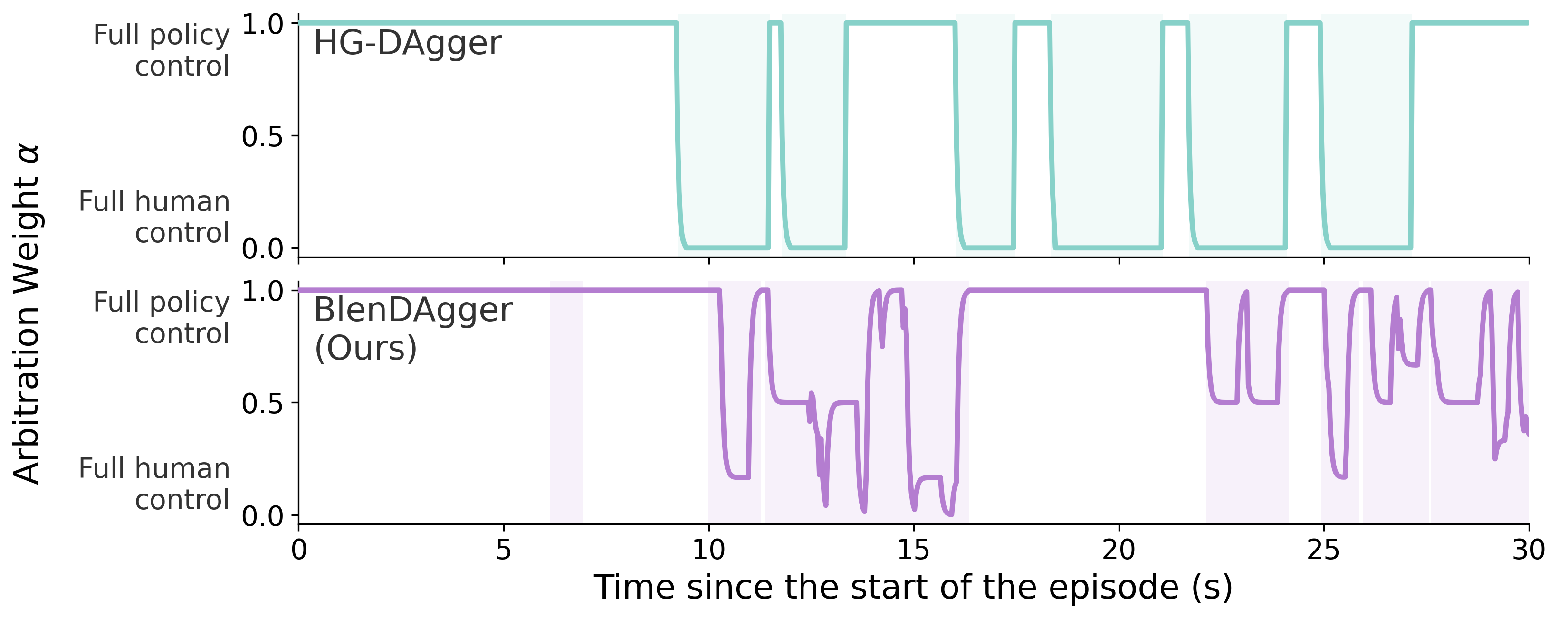}
\caption{$\alpha$ over an example 30-second segment. Highlighted sections show timesteps where the participant is intervening.}
\label{fig:alpha_over_time}
\end{figure}

We use EMA smoothing with the same value of $\lambda=0.5$ for HG-DAgger at the start of interventions. This is to provide a fair comparison and isolate the effect of blending \textit{during} interventions from the effect of smoothing. An example of $\alpha$ values over time from a segment of a participant's collection episode is shown in Fig.~\ref{fig:alpha_over_time}.

\section{Methods}
We evaluate the impact of BlenDAgger on policy performance with five manipulation tasks, three in simulation and two in the real world. We train base policies on human demonstrations and then fine-tune with interventions collected by two expert demonstrators, who are authors on this paper, with either HG-DAgger or BlenDAgger. For all tasks, we perform five iterative rounds of HG-DAgger and BlenDAgger. Per method and per round, 10 episodes are collected for the simulation tasks and 15 episodes for the real-world tasks. We use the diffusion policy architecture and training parameters described in Section~\ref{sec:arch}. The following sections describe the tasks and evaluation, and Table~\ref{tab:randomization} describes the task randomization. To evaluate user perceptions of BlenDAgger and whether non-expert data improves policy performance, we also conduct a user study (n=14)\footnote{This study was approved by an ethics review board. All participants provided informed consent.}, described in Section~\ref{sec:user_study}.

\subsection{Simulation Experiments}
We first evaluate our approach in the RoboCasa simulation environment~\cite{Nasiriany_Robocasa_2024}, using a 7DoF Franka Emika Panda robot arm. We use the \textit{PrepareCoffee}, \textit{MicrowaveThawing}, and \textit{PickPlaceCounterToCab} tasks (see Fig.~\ref{fig:tasks}(\subref{fig:coffee-task}--\subref{fig:lemon-task})), but use one fixed scene and object per task with randomized object and robot starting positions and orientations. From a base policy trained on 50 human demonstrations, we fine-tune with interventions collected using a SpaceMouse device. We train with Intervention-Weighted Regression
(IWR)~\cite{Mandlekar_IWR_2020}, drawing half of all training samples from $\mathcal{D}_{I}$, which contains only the intervention segments, and
half from $\mathcal{D}_{R}$, which contains the base demonstrations and the
autonomous rollout segments. Policy performance is evaluated as the binary success rate on the task, averaged over 100 rollouts. 

\begin{figure*}[t]
\centering
\begin{minipage}[t]{0.4\textwidth}
\centering
\begin{subfigure}[t]{0.5\linewidth}
    \centering
    \includegraphics[width=\linewidth]{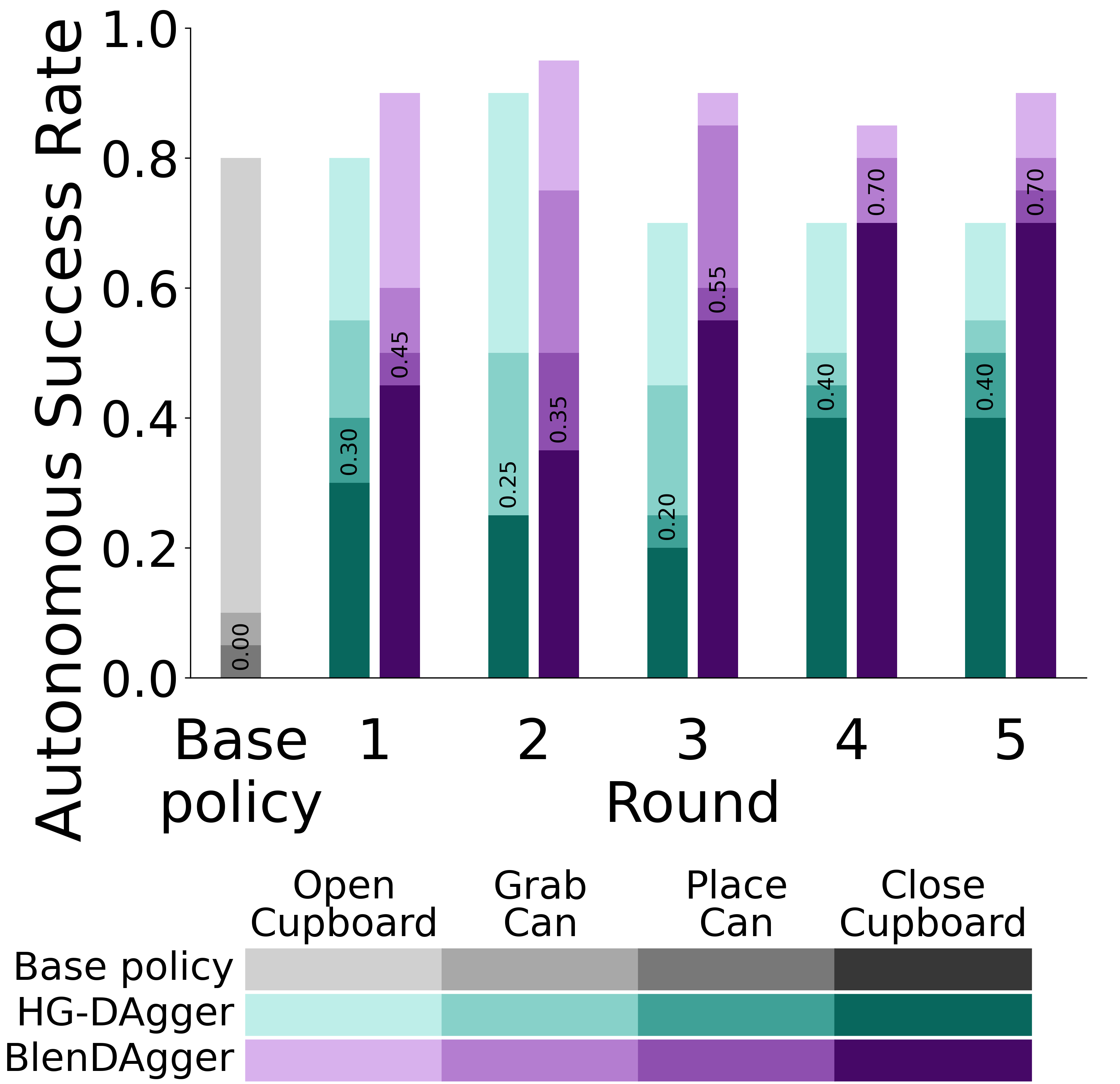}
    \caption{Cupboard Stowing}
    \label{fig:cupboard_success}
\end{subfigure}%
\begin{subfigure}[t]{0.5\linewidth}
    \centering
    \includegraphics[width=\linewidth]{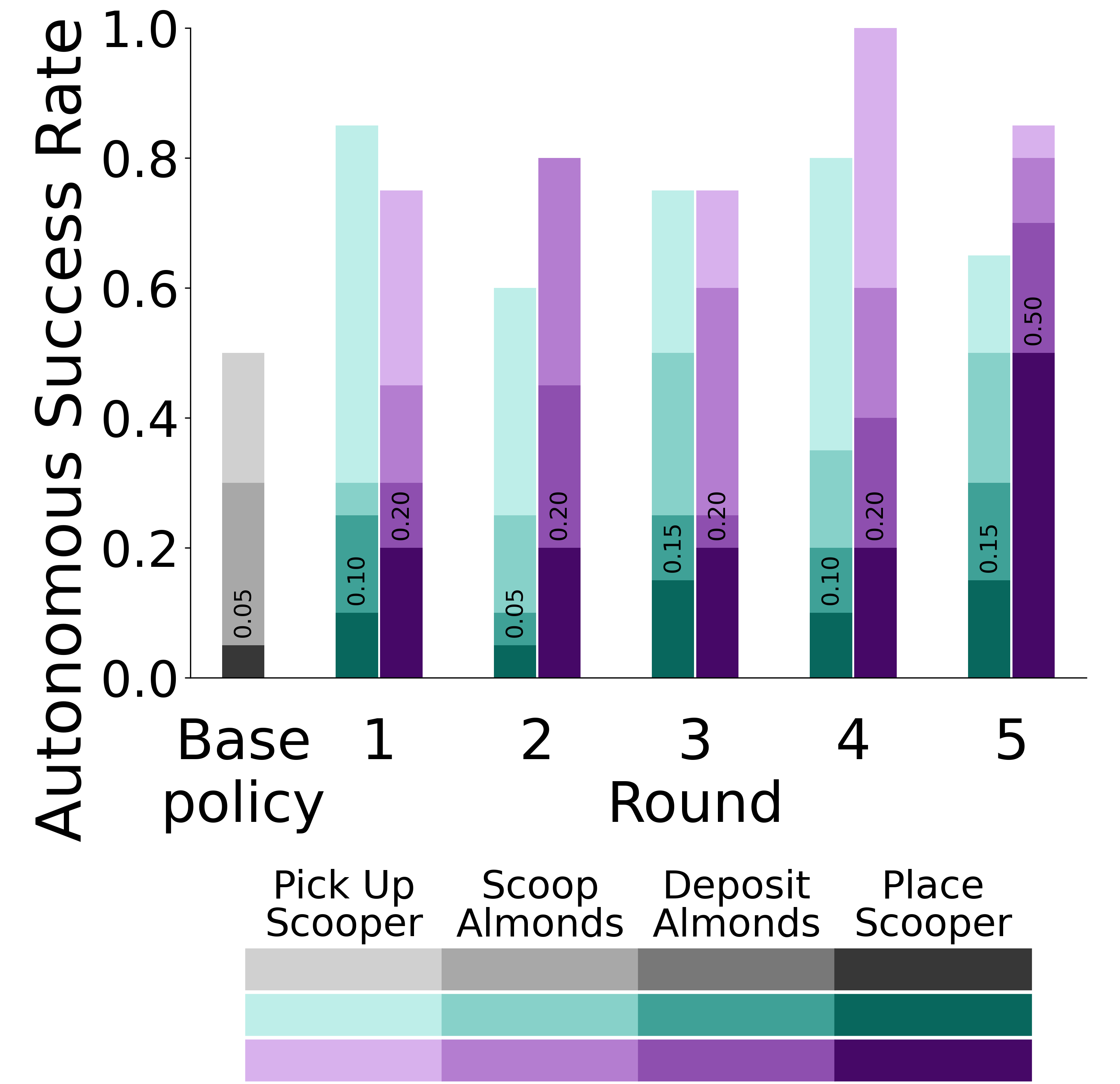}
    \caption{Almond Scooping}
    \label{fig:almond_success}
\end{subfigure}
\end{minipage}%
\begin{minipage}[t]{0.6\textwidth}
\centering
\begin{subfigure}[t]{0.3333\linewidth}
    \centering
    \includegraphics[width=\linewidth]{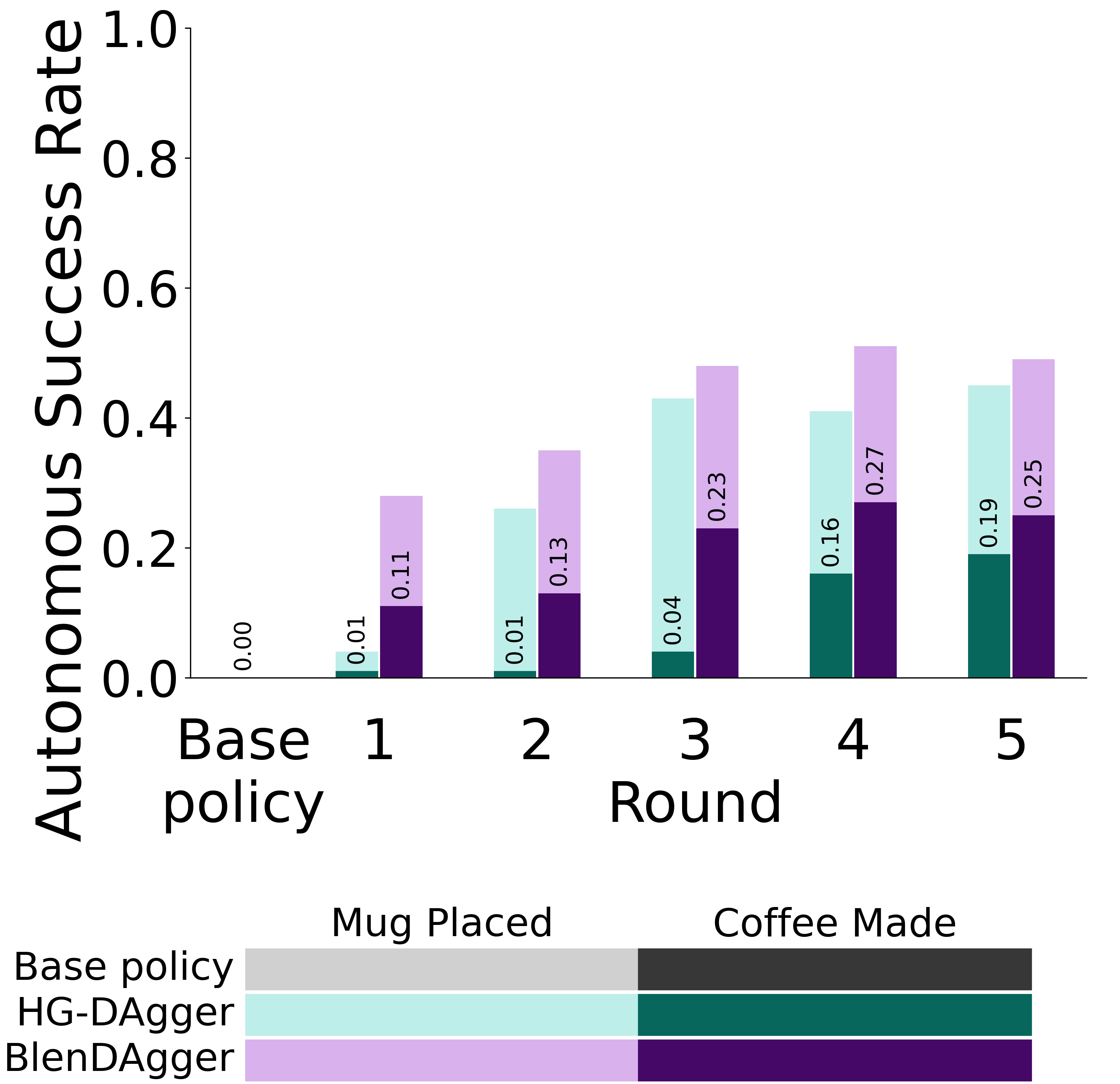}
    \caption{Prepare Coffee}
    \label{fig:coffee_success}
\end{subfigure}%
\begin{subfigure}[t]{0.3333\linewidth}
    \centering
    \includegraphics[width=\linewidth]{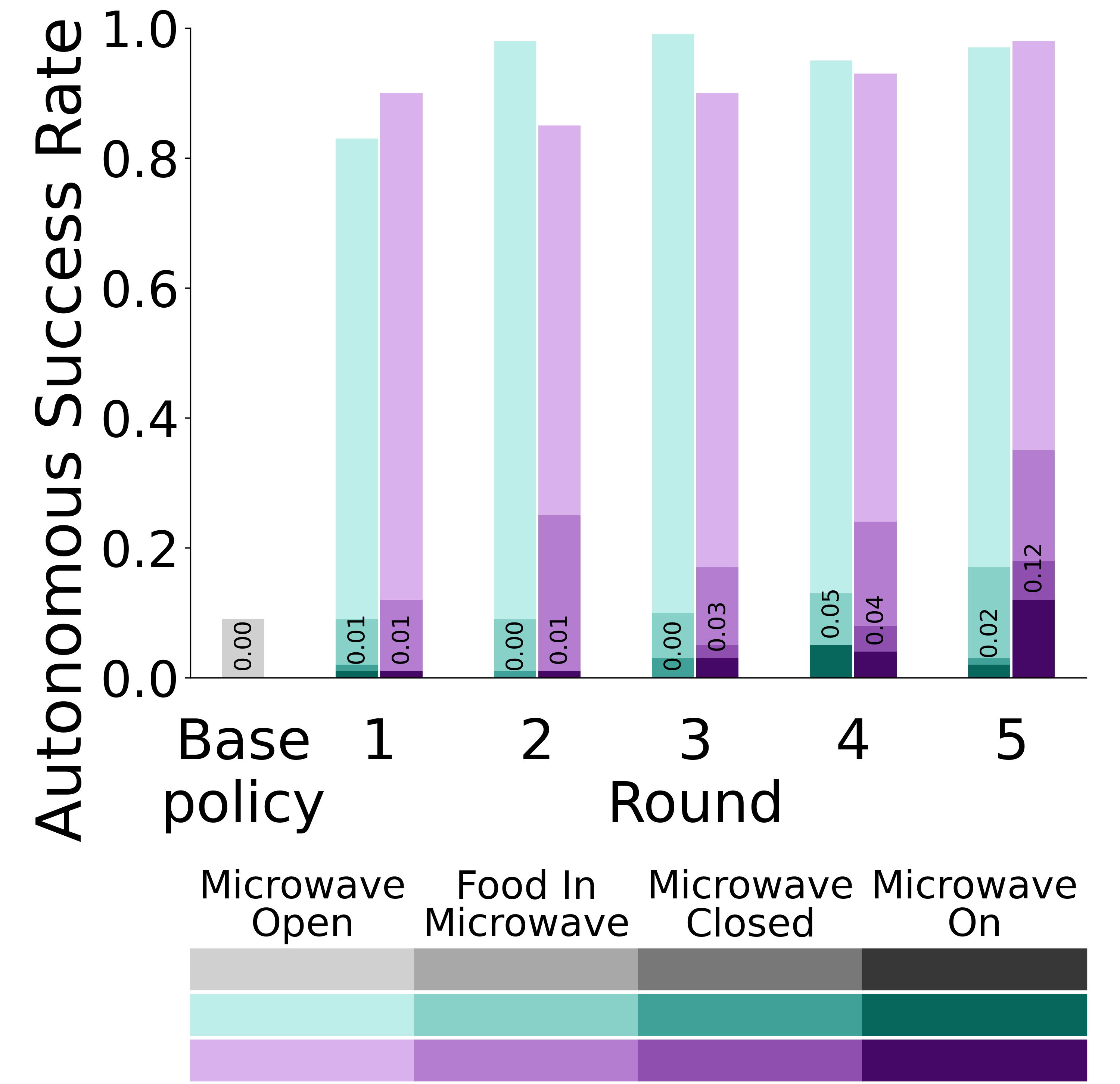}
    \caption{Microwave Thawing}
    \label{fig:thawing_success}
\end{subfigure}%
\begin{subfigure}[t]{0.3333\linewidth}
    \centering
    \includegraphics[width=\linewidth]{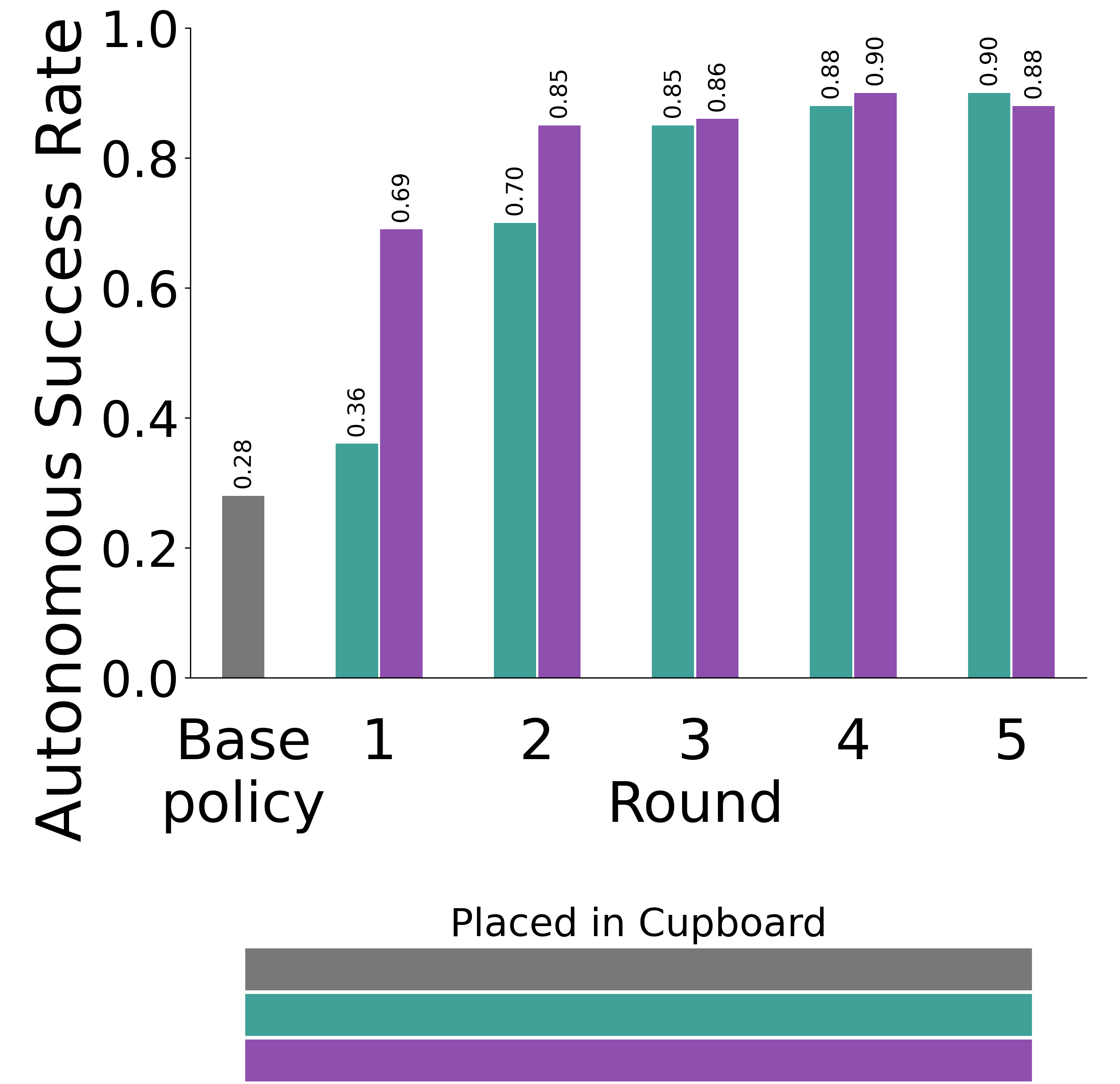}
    \caption{Pick and Place}
    \label{fig:lemon_success}
\end{subfigure}
\end{minipage}
\caption{Autonomous performance on real-world (a--b) and simulation (c--e) tasks. Bar colors indicate the highest subtask reached during an episode. Full task success is labeled.}
\label{fig:performance_real}
\label{fig:performance_sim}
\end{figure*}

\subsection{Real-World Experiments}
To validate our simulation results with real-world data collection, we perform real-world experiments with two long-horizon tasks, shown in Fig.~\ref{fig:tasks}(\subref{fig:cupboard-task}--\subref{fig:almond-task}), using a 6DoF UFACTORY xArm robot. Data was collected by trained demonstrators using an Oculus VR controller, and participants used a SpaceMouse since it can be easier to learn to use. The \textit{Cupboard Stowing} task is motivated by household grocery stowing. This task involves opening a cupboard door, picking up a can, placing the can in the cupboard, and closing the cupboard door. The \textit{Almond Scooping} task is motivated by grocery bulk-bin shopping. It requires picking up a scooper, scooping almonds, depositing almonds in a bag, and returning the scooper to a napkin. The task is considered a success if no almonds are spilled and $>$5 almonds are deposited. 

For each task, we collect 45 demonstrations to train a base policy and then fine-tune with rounds of interventions. After finding that policy performance decreased if we trained with autonomous segments where the human did not intervene, likely due to the autonomous rollouts being lower quality than in simulation, we draw half of the training samples from $\mathcal{D}_{I}$ and half from the original base policy demonstrations. Autonomous policy performance after each round of fine-tuning is averaged over 20 rollouts.

\subsection{User Study}
\label{sec:user_study}
To evaluate the experience of providing corrections with our approach, we recruited 15 participants. We restricted analysis to non-expert users, removing one participant who reported the maximum rating on both robotics-experience measures, leaving n=14. Intervention method (BlenDAgger vs. HG-DAgger) was varied within subjects and the task (the two real-world tasks) was varied between subjects. Within subject, we varied the policy the methods were applied to (base policy vs. round-1 fine-tuned policy). This yielded four trial sets per participant (2 intervention methods $\times$ 2 policies). The round-1 fine-tuned policy was fine-tuned on expert data to remain consistent across participants and due to time constraints of fine-tuning during the study. This allowed for measuring the collection experience, but did not evaluate the compounding effect of participant data across rounds.

After signing informed consent, each participant was shown how to use the SpaceMouse device and practiced using it. They were introduced to the task and practiced it twice with full teleoperation, twice with BlenDAgger, and twice with HG-DAgger. The order of intervention methods was counterbalanced between participants. Experimental trials began, in which they completed the four sets of five trials each.  Participants answered a questionnaire after each of the four sets of trials. 
The questionnaire contained the following:
\begin{enumerate}
    \item Weighted NASA TLX Workload Scale~\cite{Hart_TLX_1988}.
    \item Four Likert scales, containing three 7-point items per scale. These were: \textit{I knew when the robot needed to be corrected}, \textit{I was willing to intervene}, \textit{The robot helped me provide corrections}, and \textit{Perceived smoothness} (Cronbach's alpha $= 0.83$, $0.77$, $0.48$, $0.82$).
\end{enumerate}

Of the participants, 9 were female and 5 were male, with a mean age of 30.4 (SD=12.5). On a scale of 0 to 5 (0=no experience, 5=professional experience), the mean prior experience with robotics was 1.3 (SD=1.0). On a scale of 0 to 5 (0=no experience, 5=regularly control a robot arm), the mean prior experience controlling a robot arm was 0.5 (SD=0.7). The study lasted two hours and participants were compensated.

\subsection{Statistical Analysis}
We conducted Bayesian analysis using~\cite{Morey_BayesFactorR_2015}, which allows for reporting effects both \textit{for} and \textit{against} a hypothesis. We report $BF_{10}$ from the Bayesian equivalent of RM-ANOVAs, with method and round as repeated measures. The random effect is participant for the user study and task for the expert data. Results were interpreted using~\cite{Lee_Bayes_2014}, with BF$\in [0.333, 3.0]$ as inconclusive, BFs above 3.0 as evidence for an effect, and BFs below
0.333 as evidence against an effect. For example, a Bayes factor of 3 means the data are 3 times more likely under the alternative hypothesis than under the null. 

\subsection{Hypotheses}
We hypothesize that, compared to HG-DAgger:
\begin{itemize}
    \item \textbf{H1}: Policies trained with BlenDAgger data will achieve higher autonomous performance.
    \item \textbf{H2}: BlenDAgger will require less human effort to collect intervention data. 
    \item \textbf{H3}: BlenDAgger will yield higher subjective perceptions (measured via workload and Likert scales).
\end{itemize}


We test \textbf{H1} and \textbf{H2} in our main data-collection experiments, and further evaluate all three in a user study with non-expert operators to assess whether the benefits extend beyond trained demonstrators. We also analyze whether the performance improvements from BlenDAgger correspond with our motivations for using shared control, described in Section~\ref{sec:intro}.

\subsection{Analysis Metrics}
We call a timestep $t$ an \emph{intervention step} if $M_t > 0$, and a run of consecutive intervention steps an \emph{intervention}. We define the following metrics for later analysis, averaged within an episode:
\begin{itemize}
\item \emph{Action discontinuity} is $\lVert \pi_i(x_t) -
\pi_i(x_{t-1}) \rVert_2$ at the first and last step of each intervention. We use this as a proxy for the smoothness of a control change, where lower action discontinuity indicates greater smoothness.
\item  \emph{Intervention timing} is
the timestep of each control transition, measured from the start of the episode. Since the policies have different failure modes after fine-tuning, we only compare intervention timing on round 1 data.
\item \emph{State distance} is the mean distance, in joint-angle space, from a state to its $10$ nearest neighbors~\cite{Angiulli_KNN_2002} in the training data. The training data includes base policy demonstrations and interventions from previous rounds.
\item \emph{Trajectory distance} is the discrete Fr\'echet distance~\cite{Eiter_Frechet_1994}, in joint-angle space, from an intervention to the closest equal-length trajectory segment in the training data.
\end{itemize}

\subsection{Policy Architecture and Training Parameters}
\label{sec:arch}
Simulation and real-world experiments both use a diffusion policy~\cite{Chi_Diffusion_2024}
with a 1-D temporal convolutional U-Net and a per-camera ResNet-18 visual encoder.
In simulation, we use the Robomimic implementation~\cite{Mandlekar_Robomimic_2021}, and in the real world, we use the Diffusion Policy codebase~\cite{Chi_Diffusion_2024}.
Table~\ref{tab:policy} describes the architecture and training hyperparameters used. For each task, the base policy checkpoint was chosen by a short evaluation to determine the best-performing one. These were found to be epoch 4000 for \textit{Cupboard Stowing}, epoch 3750 for \textit{Almond Scooping}, and epoch 900 for \textit{Pick and Place}. The other two simulation tasks had 0\% task success, so we chose epoch 1000. Each fine-tuning round initializes from the base policy checkpoint and freezes the visual encoder.

\begin{table}[t!]
\centering
\scriptsize
\renewcommand{\arraystretch}{1.1}
\newcommand{\ind}{\hspace{1.5em}}
\setlength{\tabcolsep}{14pt}
\caption{Policy architecture and training hyperparameters.}
\label{tab:policy}
\begin{tabular}{@{}l@{\hspace{1.5em}}cc@{}}
\toprule
 & \textbf{Simulation} & \textbf{Real} \\
\midrule
\textbf{Architecture} & & \\
\ind External / wrist cameras   & 2 / 1 & 1 / 1 \\
\ind Image resolution           & \multicolumn{2}{c}{$128 \times 128$} \\
\ind Random crop                & $116 \times 116$ & $112 \times 112$ \\
\ind Proprioception dim.        & 16 & 14 \\
\ind Action dim.                & 12 & 10 \\
\ind Obs.\ history $T_o$        & 2 & 3 \\
\ind Pred.\ horizon $T_p$       & \multicolumn{2}{c}{16} \\
\ind Executed actions $T_a$     & \multicolumn{2}{c}{8} \\
\ind Training / DDIM inference steps & \multicolumn{2}{c}{100 / 16} \\
\midrule
\textbf{Training} & & \\
\ind Optimizer (base)           & Adam & AdamW \\
\ind Optimizer (fine-tune)       & \multicolumn{2}{c}{AdamW} \\
\ind LR (base)                  & $1{\times}10^{-4}$ & $5{\times}10^{-4}$ \\
\ind LR (fine-tune)              & \multicolumn{2}{c}{$1{\times}10^{-4}$} \\
\ind Schedule (base)            & constant & cosine \\
\ind Schedule (fine-tune)        & \multicolumn{2}{c}{constant w/ warmup} \\
\ind Batch size (base)          & 128 & 512 \\
\ind Batch size (fine-tune)      & \multicolumn{2}{c}{512} \\
\ind Epochs (fine-tune)          & 150 & 500 \\
\bottomrule
\end{tabular}
\end{table}

\begin{table}[t]
\centering
\scriptsize
\setlength{\tabcolsep}{2.5pt}
\caption{Randomization for data collection and evaluation.}
\label{tab:randomization}
\begin{tabular}{@{}llcccccc@{}}
\toprule
 & \multicolumn{3}{c}{Object} & \multicolumn{2}{c}{Robot base} & \multicolumn{2}{c}{RoboCasa} \\
\cmidrule(lr){2-4}\cmidrule(lr){5-6}\cmidrule(lr){7-8}
Task & Object & \begin{tabular}[t]{@{}c@{}}$x$/$y$ Pos.\\ (cm)\end{tabular} & \begin{tabular}[t]{@{}c@{}}Yaw\\ (rad)\end{tabular} & \begin{tabular}[t]{@{}c@{}}$x$/$y$ Pos.\\ (cm)\end{tabular} & \begin{tabular}[t]{@{}c@{}}Yaw\\ (rad)\end{tabular} & Layout & Style \\
\midrule
Prepare Coffee    & Mug       & $\pm$10 / 10 & $0$ & $\pm$2.5 / 4 & $\pm0.17$ & 7 & 9 \\
Microwave Thawing & Carrot & $\pm$5 / 5   & $0$ & $\pm$1.5 / 4 & $\pm0.17$ & 4 & 0 \\
Pick and Place    & Lemon  & $\pm$10 / 10 & $2\pi$ & $0$ & $0$ & 1 & 1 \\
\midrule
Cupboard Stowing & Can      & $\pm$11 / 11 & $2\pi$  & $0$ & $0$ & --- & --- \\
                 & Cupboard & $0$ / $0$    & $\pi/6$ & $0$ & $0$ & --- & --- \\
\midrule
Almond Scooping & Bag / Scooper & $\pm$5 / 8 & $0$ & $0$ & $0$ & --- & --- \\
                & Almond bin    & $\pm$3 / 5 & $0$ & $0$ & $0$ & --- & --- \\
\bottomrule
\end{tabular}
\end{table}

\section{Results}
\subsection{Performance Results}
As shown in Fig.~\ref{fig:performance_real}(\subref{fig:cupboard_success}--\subref{fig:almond_success}), BlenDAgger results in higher autonomous performance than fine-tuning with HG-DAgger in every round on both real tasks, ending 35 percentage points higher on the \textit{Almond Scooping} task (50\% vs.\ 15\%) and 30 percentage points higher on the \textit{Cupboard Stowing} task (70\% vs.\ 40\%). As shown in Fig.~\ref{fig:performance_sim}(\subref{fig:coffee_success}--\subref{fig:lemon_success}), BlenDAgger also achieves higher success than HG-DAgger on two of the simulated tasks: on \textit{Prepare Coffee} it is higher in every round (reaching 27\% vs.\ 19\%), and on \textit{Microwave Thawing} it reaches 12\% vs.\ 5\%, which is low in absolute terms across both methods due to the task being long-horizon with 0\% base policy performance. On \textit{Pick and Place}, BlenDAgger achieves 85\% task success with 33\% fewer episodes of interventions than HG-DAgger, but both flatline at  $\sim$90\% success, demonstrating that BlenDAgger's main performance gains are for complex, long-horizon tasks. The results on these five tasks support \textbf{H1} for expert-collected data. BlenDAgger reduces data collection time by 5.9\% (BF=70.8), while the number of intervention steps and total magnitude of interventions do not differ reliably between methods (BF=0.57, 0.28). This supports \textbf{H2} for expert demonstrators. Since the number of intervention steps and magnitude of interventions do not differ significantly, the performance gap is attributable to blended control, rather than when or how the demonstrator chose to intervene.

\begin{figure}[t]
\centering
\begin{subfigure}[b]{0.24\textwidth}
    \centering
    \includegraphics[width=\textwidth]{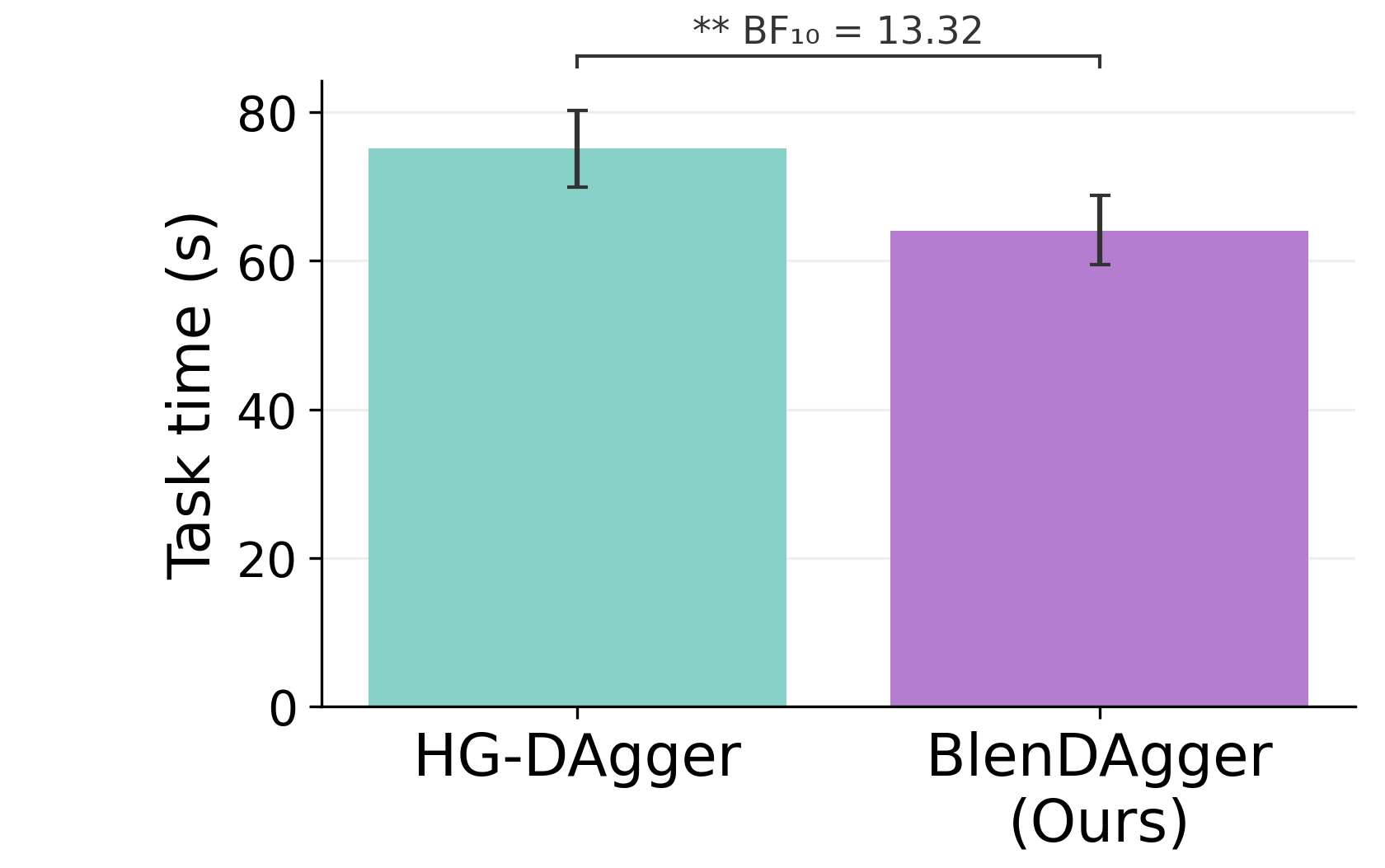}
    \caption{Data Collection Time}
    \label{fig:collection_time}
\end{subfigure}
\hfill
\begin{subfigure}[b]{0.24\textwidth}
    \centering
    \includegraphics[width=\textwidth]{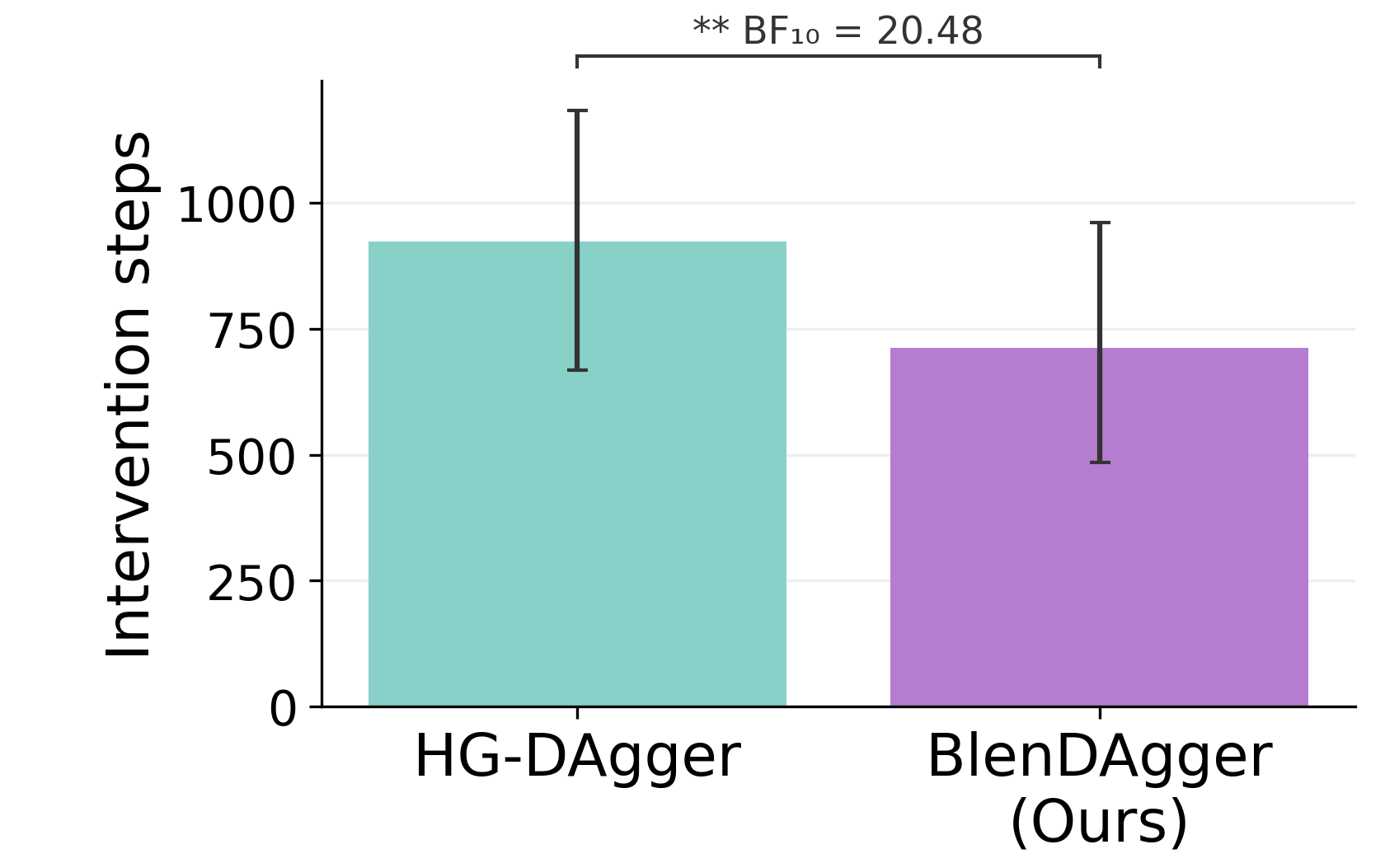}
    \caption{Intervention Steps}
    \label{fig:intervention_steps}
\end{subfigure}
\caption{Participants complete the task faster and with fewer intervention steps with BlenDAgger. Error bars show 95\% CIs.}
\label{fig:data_collection}
\end{figure}

\begin{figure}[t]
\centering
\begin{subfigure}[b]{0.24\textwidth}
    \centering
    \includegraphics[width=\textwidth]{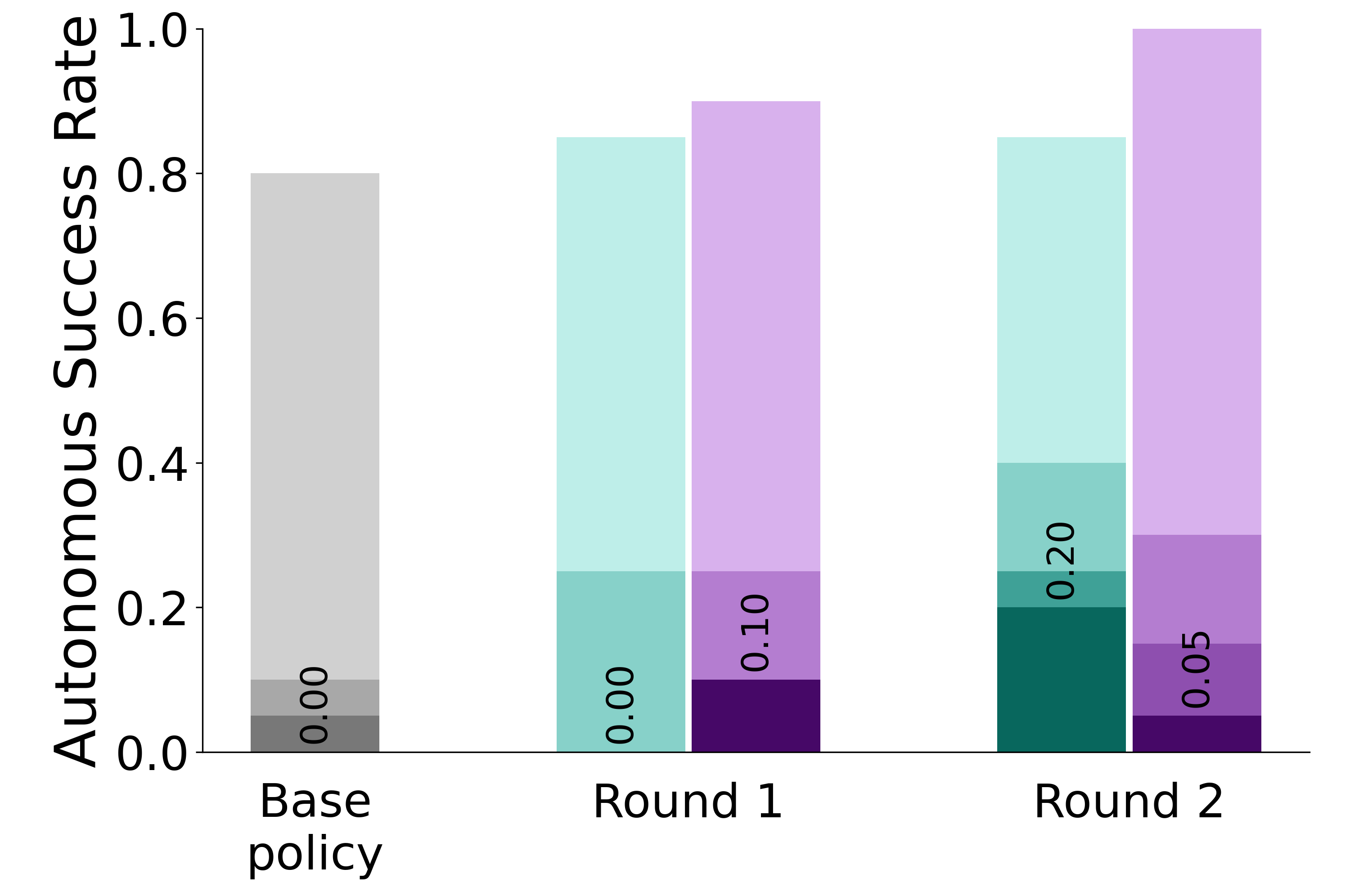}
    \caption{Cupboard Stowing}
    \label{fig:cupboard_user_study}
\end{subfigure}
\hfill
\begin{subfigure}[b]{0.24\textwidth}
    \centering
    \includegraphics[width=\textwidth]{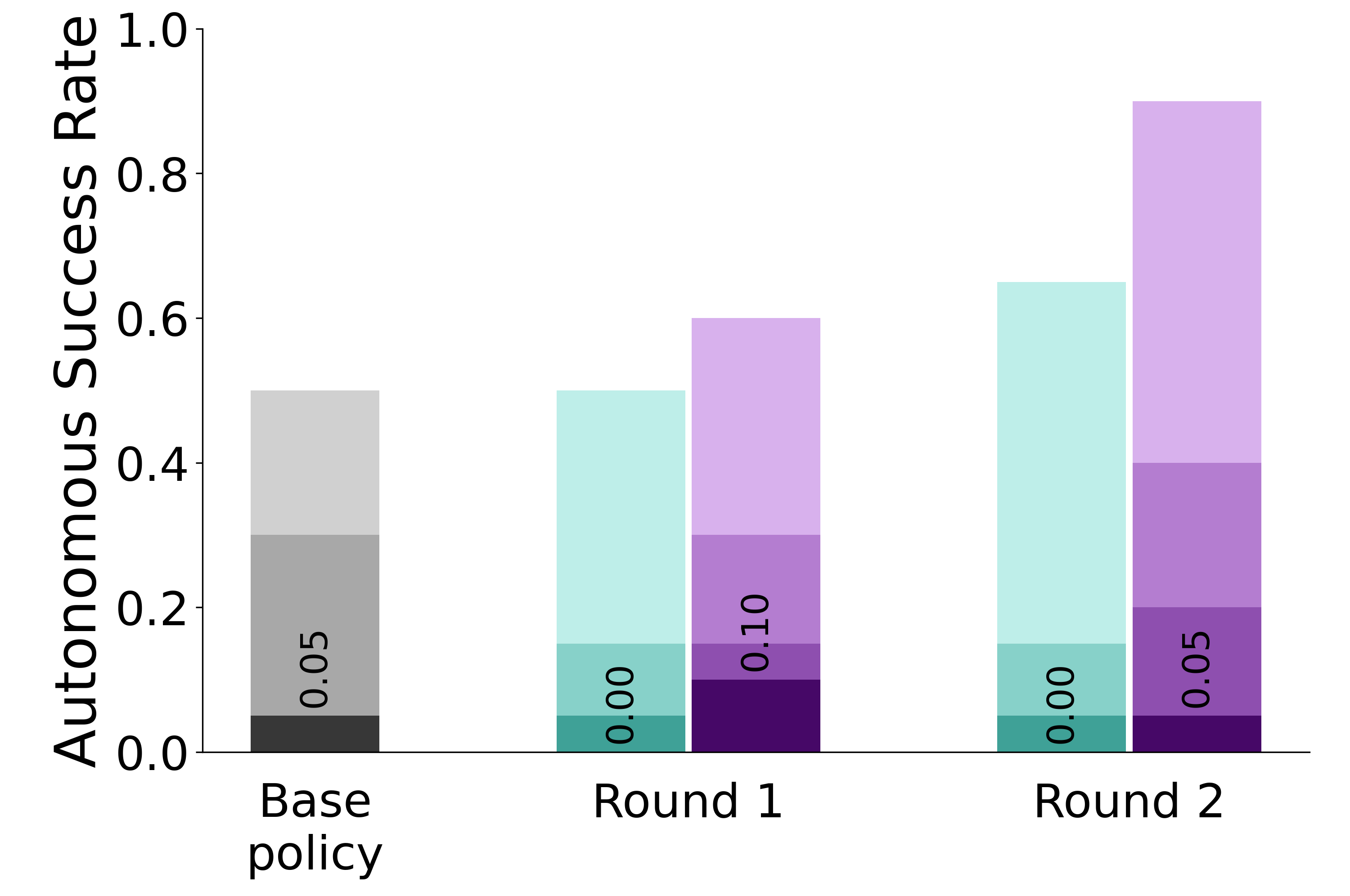}
    \caption{Almond Scooping}
    \label{fig:almond_user_study}
\end{subfigure}
\caption{Autonomous performance (user study data).}
\label{fig:user_study_performance}
\end{figure}


\subsection{User Study Results}
In a user study, BlenDAgger resulted in lower task completion time on successful episodes (BF=13.32), shown in Fig.~\ref{fig:data_collection}. BlenDAgger also resulted in fewer intervention steps (BF=20.48), using a rank-based test because the paired differences were strongly skewed (Shapiro–Wilk $p=.001$). We did not find statistically significant differences in task success during data collection (BF=0.63), workload (BF=0.54), or three of the Likert scales (\textit{Knew when to correct} BF=0.43, \textit{Willing to intervene} BF=0.60, \textit{Helped during corrections} BF=0.63). The \textit{Helped during corrections} scale had low internal consistency (Cronbach's alpha = 0.48), so its result should be interpreted with caution. There was evidence \textit{against} an effect on the \textit{Perceived smoothness} Likert scale (BF=0.26). These results support \textbf{H2} for participant data collection but do not support \textbf{H3}, showing that BlenDAgger reduces task completion time and time spent intervening but does not result in an effect on subjective perceptions. 


As shown in Fig.~\ref{fig:user_study_performance}, participant-collected BlenDAgger data produced policy improvement on both rounds of both tasks and outperformed HG-DAgger in three of four conditions: both rounds of \textit{Almond Scooping} and round 1 of \textit{Cupboard Stowing}. On the \textit{Almond Scooping} task, BlenDAgger increased performance in both rounds, whereas HG-DAgger caused performance to decrease to 0\% full task success in both rounds. Consistent with prior findings that low-quality interventions limit policy performance~\cite{Mandlekar_Robomimic_2021, Sakr_DataConsistency_2025}, there remains a performance gap between data collected by trained demonstrators and non-experts. This provides some support for \textbf{H1} on user study data, but shows that limitations remain with training on robot manipulation data collected by novice users.
\subsection{Analysis of BlenDAgger Advantages}
\begin{figure}[t!]
\centering
\begin{subfigure}[b]{0.49\textwidth}
    \centering
    \includegraphics[width=\textwidth]{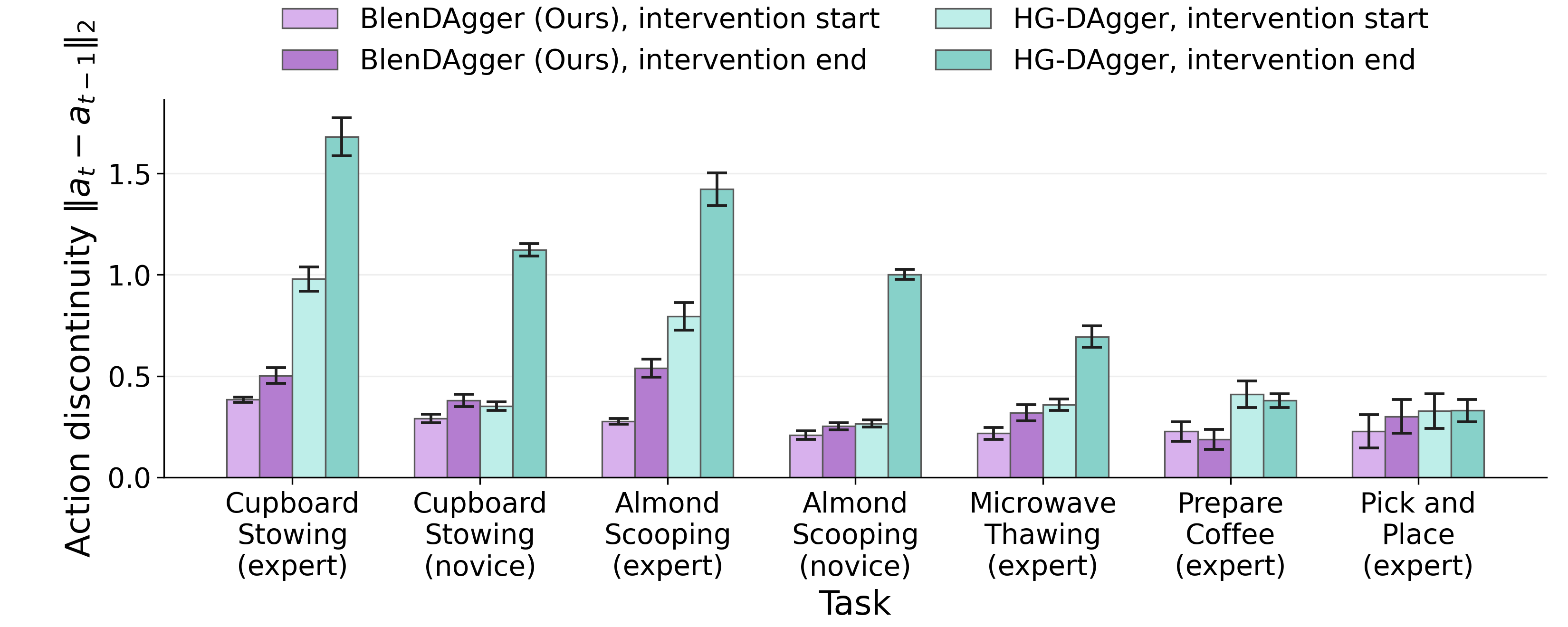}
    \caption{Action Discontinuity during Control Transitions}
    \label{fig:smooth_analysis}
\end{subfigure}
\hfill
\begin{subfigure}[b]{0.49\textwidth}
    \centering
    \includegraphics[width=\textwidth]{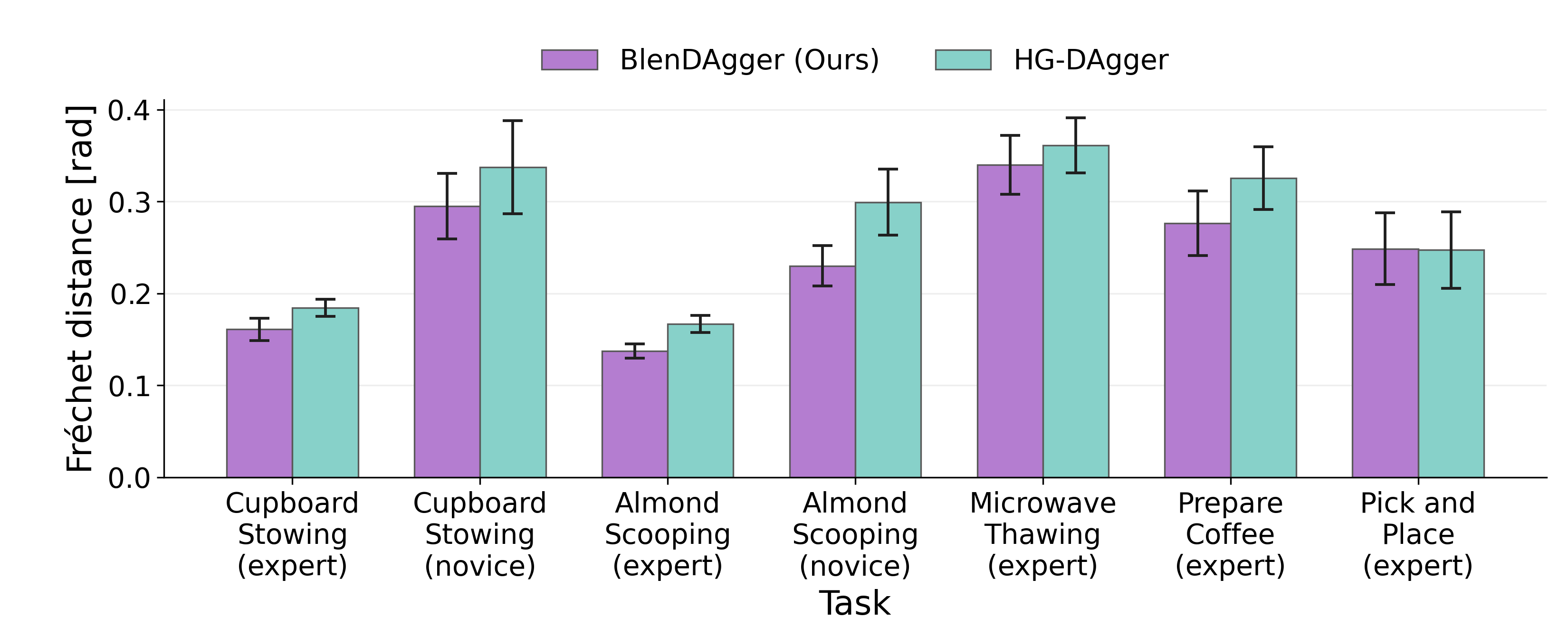}
    \caption{Trajectory Distance to Training Data}
    \label{fig:traj_analysis}
\end{subfigure}
\caption{BlenDAgger has smoother action transitions and lower trajectory distance to training data. Error bars show 95\% CIs.}
\label{fig:blended_analysis}
\end{figure}
Based on the performance improvements supporting \textbf{H1}, we analyze the advantages of BlenDAgger and report the BFs in Table~\ref{tab:bayes_factors}. As shown in Fig.~\ref{fig:blended_analysis}, we find that BlenDAgger results in 57\% smoother control transitions when interventions start and end compared to HG-DAgger. Critically, these smoother transitions are caused by blending, not by smoothing at control changes, since HG-DAgger and BlenDAgger share an EMA smoothing value of $0.5$. During interventions, the mean arbitration was $\alpha=0.50$ (SD 0.32) for expert demonstrators and $\alpha=0.57$ (SD 0.33) for participants, confirming that control was genuinely shared and that $\alpha$ was similar between trained demonstrators and participants. For novices, we find that BlenDAgger visits fewer highly OOD states (${>}99$\textsuperscript{th} percentile of training distances) than HG-DAgger. We further find that BlenDAgger interventions have 14\% higher trajectory similarity to the training data. The timing of interventions is inconclusive. This analysis suggests that higher smoothness and higher trajectory similarity to training data may contribute to BlenDAgger's policy performance improvements. 

\begin{table}[t]
\centering
\scriptsize
\caption{BFs comparing BlenDAgger to HG-DAgger on analysis metrics. Bold indicates evidence of an effect.}
\label{tab:bayes_factors}
\begin{tabular}{lcc}
\toprule
Metric & Experts & Participants \\
\midrule
Action discontinuity, intervention start/end & {\boldmath$6.2\mathrm{e}{7}$} / {\boldmath$1.5\mathrm{e}{6}$} & {\boldmath$15.34$} / {\boldmath$3.8\mathrm{e}{36}$} \\
Highly OOD states visited     & $0.34$ & {\boldmath$17.07$} \\
Trajectory distance to training data   & {\boldmath$3.35$} & {\boldmath$3.10$} \\
Intervention timing, intervention start/end   & $0.51$ / $0.53$ & $0.99$ / $1.66$ \\
\bottomrule
\end{tabular}
\end{table}

\subsection{Ablations}
\label{sec:ablations}
We have two desired aspects for a blending formulation: 1) increased policy performance and 2) sufficient control over the robot when providing corrections. On the \textit{Prepare Coffee} simulation task, we compare our BlenDAgger formulation against a fixed arbitration of $\alpha=0.5$ and adaptive blending ablations with each of the three components described in Section~\ref{sec:arbitration} removed and each on its own. As shown in Table~\ref{tab:ablation_success}, we find that many of the ablations result in similar autonomous performance, demonstrating that blended shared control is the primary cause of higher performance, rather than a specific arbitration. These results also indicate that the BlenDAgger arbitration is not adapted specifically to our demonstrators as performance gains remain in a fixed-$\alpha$ approach. The components in BlenDAgger instead contribute towards the demonstrator's ability to quickly regain some control of the robot's movement, which we measure as the median time from the onset of an intervention to when the robot's executed action is within 20\% of the user's commanded action direction and speed, shown in Table~\ref{tab:ablation_success}. We evaluate the effect of executing blended actions while training only on the portion corresponding to human actions rather than blended actions (\textit{Human action targets}), which results in lower autonomous task success. We also show the mean and standard deviation in performance of HG-DAgger and BlenDAgger across three training seeds; the first seed is used for the following round of data collection.


\begin{table}[t]
\centering
\scriptsize
\caption{Ablation performance and responsiveness. Bold marks the best overall and the best BlenDAgger variant.}
\label{tab:ablation_success}
\setlength{\tabcolsep}{4pt}
\begin{tabular}{@{}lccc@{\hspace{14pt}}c@{}}
\toprule
 & \multicolumn{3}{c}{\textbf{Performance}} & \raisebox{-1.1ex}[0pt][0pt]{\textbf{Responsive-}} \\
\cmidrule(lr){2-4}
\textbf{Method} & \textbf{R1} & \textbf{R2} & \textbf{R3} & \textbf{ness (s)} \\
\midrule
\textbf{HG-DAgger} & 0.01\,\textpm\,0.00 & 0.02\,\textpm\,0.01 & 0.05\,\textpm\,0.01 & \textbf{0.15} \\
\textbf{BlenDAgger} & \textbf{0.08\,\textpm\,0.03} & 0.15\,\textpm\,0.02 & \textbf{0.27\,\textpm\,0.04} & \textbf{0.20} \\
\hspace{0.6em}Without uncertainty & 0.00 & 0.14 & 0.16 & 0.28 \\
\hspace{0.6em}Without similarity & 0.01 & 0.15 & 0.22 & \textbf{0.20} \\
\hspace{0.6em}Without magnitude & 0.03 & 0.13 & 0.17 & \textbf{0.20} \\
\hspace{0.6em}Only uncertainty & \textbf{0.08} & 0.09 & 0.23 & 0.33 \\
\hspace{0.6em}Only similarity & 0.00 & 0.15 & 0.18 & 0.85 \\
\hspace{0.6em}Only magnitude & 0.01 & \textbf{0.18} & 0.11 & 1.02 \\
\hspace{0.6em}Fixed-$\alpha$ ($\alpha=0.5$) & 0.03 & 0.13 & 0.15 & 1.15 \\
\hspace{0.6em}Human action targets & 0.00 & 0.05 & 0.23 & --- \\
\bottomrule
\end{tabular}
\end{table}

\section{Limitations and Future Work}
In this paper, we show simulation and real-world results for blended shared control as an alternative to HG-DAgger for learning manipulation tasks. While our arbitration formulation is inspired by prior shared-control literature and validated through ablations, we adopt fixed arbitration parameters and DoF groupings. Future work could learn or adapt these online, potentially personalizing to each operator. Although intervention data in interactive imitation learning is typically collected by a small number of skilled operators, frequently the researchers themselves~\cite{Kelly_HG-DAgger, Mandlekar_IWR_2020, Liu_Learning-on-the-job}, validating BlenDAgger with a larger pool of trained operators could help characterize how robustly the approach performs. Our ablations (Section~\ref{sec:ablations}) suggest the performance gains do not depend on demonstrators adapting to the arbitration, as performance persists under different arbitrations, so we would expect to see similar performance trends with other demonstrators. Our user study was a single session, leaving open how users may adapt to the system over repeated use. Finally, both methods show limited gains when fine-tuned on non-expert data. Improving policy learning from imperfect demonstrations is an active area of research, and integrating blended shared control with existing approaches that filter or reweight low-quality corrections could lead to improved learning efficiency.

\section{Conclusion}
\label{sec:conclusion}
In this work, we formulate and validate a novel approach for interactive imitation learning using blended shared control, which we term BlenDAgger. BlenDAgger blends human corrections with the robot policy based on action similarity, policy uncertainty, and the amount of human intervention. Compared to HG-DAgger, our method achieves higher autonomous performance on four tasks, outperforming HG-DAgger by 30 or more percentage points on both real-world tasks. BlenDAgger also results in smoother intervention transitions and more similar trajectories to its training data. In a user study, we find statistically significant evidence that BlenDAgger speeds up data collection and do not find differences in subjective perceptions. Our results show the efficacy of blended shared control for fine-tuning manipulation policies.
\section*{Acknowledgment}
The authors thank David Yi for his support with data collection. This research is supported by ARPA-H through the University of Pittsburgh under the RAMMP (Robotic Assistive Mobility and Manipulation Platform Providing Independence for People with Disabilities) project (grant number: 75N99223S0001). This material is based upon work supported by the National Science Foundation Graduate Research Fellowship Program under Grant No(s) DGE2140739 and DGE2631988. Any opinions, findings, and conclusions or recommendations expressed in this material are those of the authors and do not necessarily reflect the views of the National Science Foundation. 

\bibliographystyle{IEEEtran}
\bibliography{references}  

\end{document}